\documentclass{article}
\usepackage{lineno,hyperref}
\modulolinenumbers[5]
\usepackage{color}
\usepackage{multirow}
\usepackage{pgfplots}
\usepackage{caption}
\usepackage{float}
\usepackage{colortbl}
\usepackage{amsmath}
\usepackage{amssymb}
\usepackage{graphicx}
\usepackage[bottom]{footmisc}
\usepackage{cleveref}
\usepackage{pst-plot}
\usetikzlibrary{babel}
\usepackage{refcount}
\usepackage{mathtools}
\usepackage{afterpage}
\usepackage{subcaption}
\usepackage{scrextend}
\usepackage{authblk}
\usepackage[english]{babel}
\usepackage[T1]{fontenc}
\pgfplotsset{compat=1.5}
\pgfplotsset
{
	width=0.5\textwidth,
	x tick label style={/pgf/number format/1000 sep=},
	enlarge x limits = 0.0,
	ymajorgrids=true,
	major tick style={draw=none},
	ymin = 0.0,
	every axis/.append style={
		every x tick label/.append style={font=\tiny},
		every y tick label/.append style={font=\tiny},
		every axis label/.append style={font=\small},
		height=37mm,
    	width=37mm,
		title style={at={(0.5,0.90)}, font=\normalfont},
		xticklabel style={yshift=4pt}
	}
}
\usepackage{ulem} %dla przekreslania tekstow

\date{}

\begin{document}

\title{Hierarchical Data Generator based on Tree-Structured Stick Breaking Process for Benchmarking Clustering Methods}
\author[1,2]{\L{}ukasz P. Olech}
\author[1]{Micha\l{} Spytkowski}
\author[1]{Halina Kwa\'snicka}
\author[2,3,4]{Zbigniew Michalewicz}

\affil[1]{Department of Computational Intelligence \protect\\ Wroc\l{}aw University of Science and Technology \protect\\ wybrze\.ze Stanis\l{}awa Wyspia\'nskiego 27, 50-370 Wroc\l{}aw, Poland\protect\\
\texttt{\{lukasz.olech,michal.spytkowski,halina.kwasnicka\}@pwr.edu.pl}}
\affil[2]{Complexica Pty Ltd. \protect\\ 155 Brebner Drive, West Lakes, SA 5021, Australia \protect\\ \texttt{\{lo,zm\}@complexica.com}}
\affil[3]{Institute of Computer Science, Polish Academy of Sciences \protect\\ ul. Ordona 21, 01-237 Warsaw, Poland}
\affil[4]{Polish-Japanese Academy of Information Technology \protect\\ ul. Koszykowa 86, 02-008 Warsaw, Poland}

\maketitle

\begin{abstract}
	Object Cluster Hierarchies is a new variant of Hierarchical Cluster Analysis that gains interest in the field of Machine Learning. Being still at an early stage of development, the lack of tools for systematic analysis of Object Cluster Hierarchies inhibits its further improvement. In this paper we address this issue by proposing a generator of synthetic hierarchical data that can be used for benchmarking Object Cluster Hierarchy methods. The article presents a thorough empirical and theoretical analysis of the generator and provides guidance on how to control its parameters. Conducted experiments show the usefulness of the data generator that is capable of producing a wide range of differently structured data. Further, benchmarking datasets that mirror the most common types of hierarchies are generated and made available to the public, together with the developed generator (\url{http://kio.pwr.edu.pl/?page\_id=396}). 
\end{abstract}

\textbf{keywords:} Artificial Data, Benchmark Data, Benchmark Data Generator, Hierarchical Clustering, Object Cluster Hierarchy, Tree-Structured Stick Breaking Process, Clustering Evaluation, Cluster Analysis.

\section{Introduction}
\label{sec:intro}
A high volume of digital data drives rapid development of analysis methods providing more effective tools to obtain data insights. The key data analysis areas of research are \textit{regression}, \textit{classification}, and \textit{clustering}~\cite{CENA2020324, 10.1007/978-981-10-8055-5_44,10.1371/journal.pone.0210236,10.1371/journal.pone.0210314, Jiang:2019:DPR:3309769.3231738, Kakkar2014}.

Clustering aims at generating \textit{meaningful} groups (clusters) of points in the provided dataset. The measure of meaningfulness of groups depends on the application and is problem-specific~\cite{deVries2019}. However, the general idea behind any measure of meaningfulness is that points within any particular cluster are \textit{as similar as possible} to each other, and at the same time, points that belong to different clusters are \textit{as different as possible} from each other. 

The clustering methods can be characterised, among others, by the type of results they produce. One of the classic types is Flat Clustering that organises points into a predefined number of clusters where the only relation between clusters is spacial. The most popular flat clustering method is k-means~\cite{TIRNAUCA201879,IJASEIT3490}, where the number of clusters is defined by one of the input parameters. Another clustering type, called \textit{Hierarchical Clustering} (\textit{HC}) or \textit{Hierarchical Cluster Analysis} (\textit{HCA}), produces several flat clustering solutions that are organised into a tree structure called dendrogram (a diagram representing a tree). In the dendrogram every node represents a cluster, and the tree structure indicates parent-child relations between clusters which don't exist in flat clustering. To obtain a flat clustering from a dendrogram, it has to be cut. The final (flat) solution can have a different number of clusters depending on where (at which level of the dendrogram) the cut is performed. The obtained flat clustering always cover all the observations from the input set as no data points are left in parent nodes. Hierarchical Clustering methods do not need to specify the number of clusters beforehand; this number is determined by the height where the dendrogram is cut. Since dendrograms have a partial order relation (hierarchical relation) between nodes, they provide insights into how the process of clustering was performed. By comparing different cuts of the same dendrogram, it can also give a view on the clustering from the perspective of different granularities (different number of clusters in the flat solution). One of the examples of hierarchical clustering is Agglomerative Hierarchical Clustering~\cite{Cohen-addad:2019:HCO:3338848.3321386,CAI2020173}.

For a variety of reasons, clustering methods, in general, still have some weaknesses~\cite{blundell2010discovering}. From the perspective of this article the interest is in the weaknesses of hierarchical methods where the primary issue is \textit{a semantic gap} between how humans perceive hierarchies and results produced by Hierarchical Clustering methods. Human perception describes hierarchical data (e.g.,~\cite{ILSVRC15}) as possessing the following properties:
\begin{enumerate}
    \item the data can be present in any node in the hierarchy and belongs to that node without being propagated to the child clusters; and
    \item the data in the child groups should represent equal or more precise concepts than the data in the corresponding parent that should resemble more general concepts; and
    \item the data in a node should be more similar to data in parent and child nodes than to unrelated nodes located in other subtrees of the hierarchy.
\end{enumerate}

\textit{Object Cluster Hierarchy} (\textit{OCH})~\cite{Spytkowski2012,Olech2016,Spytkowski2016} is a prospective extension to Hierarchical Clustering (HC) paradigm~\cite{hennig2015handbook} that aims at satisfying the above three properties, and it is further described in~\Cref{sec:intro_to_och}.

Even though the above-mentioned properties might be challenging to address, the first point has been (at least partially) incorporated in a few already-published methods such as Tree-Structured Stick Breaking for Hierarchical Clustering (TSSB-HC)~\cite{ghahramani2010tree}, Bayesian Rose Trees~\cite{2010_bayesian_rose_trees}, Inheritance Retention Variance Hierarchical Clustering (IRV-HC)~\cite{Spytkowski2012}, or modified hierarchic Gaussian Mixture Model (Hk++)~\cite{Olech2016}. Even though OCH is at an early stage of development, there are increasingly more new Object Cluster Hierarchy clustering methods being published. For that reason there is a need to establish a systematic benchmarking approach for OCH. The existing benchmark datasets created to validate HC methods are not suitable to fully validate OCH methods as they don't consider the differences between OCH and HC structures. Because of that, there is no publicly available set of benchmarking data which would assist OCH development.

This lack of commonly accessible benchmark datasets is addressed in this paper. The main contribution of the paper is the development of \textit{a new method} generating hierarchical structures of data with assumed, user-defined properties. The additional benefit for researchers is the establishment of \textit{a new set of benchmarks} --- hierarchical structures of data with the ground truth assignment. The implemented generator, together with the benchmarking datasets are freely available online at \url{http://kio.pwr.edu.pl/?page\_id=396} along with instructions on how to use it.

The published datasets can serve as a baseline benchmark for methods generating OCH. Additionally, researches can generate new structures of data according to their needs. Furthermore, such data and generator can help with proposing and testing new clustering quality measures. All these applications should significantly boost the research on OCH.

In the article, the term Object Cluster Hierarchy or its abbreviation (OCH) appears in a variety of different contexts which are best explained when compared to the classical Hierarchical Clustering. The term Hierarchical Clustering represents the \textit{concept} (approach), and similarly, Object Cluster Hierarchy represents \textit{the new concept} (paradigm). Furthermore, a dendrogram is a \textit{result} of a Hierarchical Clustering \textit{method} (e.g., hierarchical agglomerative clustering). In the case of Object Cluster Hierarchy, the \textit{result} is called an Object Cluster Hierarchy as well, or more explicitly an Object Cluster Hierarchy structure/result. A \textit{method} is referred to as Object Cluster Hierarchy method or Object Cluster Hierarchy generation method. Thus, in this article, Object Cluster Hierarchy denotes both the approach and the clustering result. The exact meaning is clear, depending on the context of use.

The paper is organised as follows. The next section provides a literature review, followed by a more in-depth introduction to the OCH in \Cref{sec:intro_to_och}. \Cref{model}  presents the details of the generator. In~\Cref{parameter}, the meaning of generator parameters and their influence on generated data are described; this section also provides insights on how to control these parameters. Conducted experiments and their results are discussed in~\Cref{experiments}, whereas, the published benchmarking datasets are presented in~\Cref{sec:benchmarking_dataset}. \Cref{conclusion} concludes the paper.

\section{Overview of benchmarking in the analysis of clustering methods}
\label{sec:review}
Clustering methods can be verified on real and/or synthetic data. The first type of data offers the advantage of representing real-world cases. However, such data may not always be available in sufficient quantities nor in a form that facilitates testing. Then, the second type of data can be used. Artificially generated data should have properties that imitate real data. A generator that produces artificial data has an additional advantage of allowing for finer control over the data used in testing, i.e., specific aspects of clustering can be tested independently.

Any new clustering method should be carefully evaluated and such evaluations are often based on a comparison to other methods.  Usually, such a comparison is made by running different methods on a number of commonly used benchmark datasets and then by comparing their results using a set of evaluation measures, e.g.,~\cite{Olech2016, Myszkowski2015, Jo2019, CAO2019185, Lakshmanaprabu2019, GARCIA2019646, WANG201950, Douglas2011544,Kakkar2014}. Benchmark dataset can be made available to the public by researchers (from academia) or come from different companies (from industry). Additionally, there are many publicly available repositories which collect and provide multiple different datasets in one place, that can be used as benchmarking baseline for method comparison, e.g., UCI repository~\cite{Dua2017UCI}, KEEL dataset repository~\cite{alcala2011keel}, and many others (e.g., \cite{2015ClusteringDatasets, OpenML2013}). 

For example, Adams et al.~\cite{ghahramani2010tree} proposed a method called TSSB-HC and tested it on two datasets -- the CIFAR-100 image set~\cite{krizhevsky2009learning}  and a sample of 1,740 documents from the NIPS 1-12\footnote{\url{https://cs.nyu.edu/~roweis/data.html}} datasets. Still, both of them lack hierarchical structure annotations (even though a hierarchy might exist within the data) so that the ability to establish an OCH structure has not been verified. Spytkowski et al.~\cite{Spytkowski2012} have proposed an extension of that method called IRV-HC. A comparison with the base method was made using a few synthetic benchmarks generated from a stochastic model of known parameters. Two measures were used: internal (Average Mixture Model Likelihood) and external (Class Purity), but again the class inclusion hierarchy was not considered. Blundell et al.~\cite{2010_bayesian_rose_trees} used several datasets to present performance of their hierarchical method Bayesian Rose Trees. One of them was a synthetically generated dataset in the form of binary vectors which is not the main focus of OCH as the vast majority of data are numerical. The other was Spambase Dataset from UCI repository -- a subset of the CMU newsgroup dataset reduced to 4 categories. The authors also used the CEDAR Buffalo\footnote{\url{http://www.cedar.buffalo.edu/Databases/}} digits dataset in two versions for testing -- a subset of the full dataset and a sample of only the 0, 2 and 4 digits. All of these datasets were used in the same way as they would be for a flat clustering algorithm, ignoring the unique capabilities of the method. In~\cite{Olech2016}, a GMM-based Hierarchical Clustering method called Hk++ was presented. UCI repository datasets~\cite{Dua2017UCI} were used for its verification, including the Iris, Wine, Glass Identification and Image Segmentation datasets. However, these datasets are not annotated with the inclusion of the hierarchical relation between classes, so this aspect of OCH couldn't be verified. In that case, elementary, synthetic and hierarchical datasets were also used, but with the purpose to show the concept rather than thoroughly verify the method capabilities. 

The above publications are the closest to the concept of Object Cluster Hierarchy that we were able to find in the literature. However, the use of data that is not annotated with a hierarchy of classes does not allow authors to verify results with respect to the obtained hierarchical structure. In the case of Rose Trees~\cite{2010_bayesian_rose_trees} in particular, the aspect was omitted due to the very conservative approach towards testing method results. In the case of TSSB method~\cite{ghahramani2010tree}, the hierarchy was thoroughly examined empirically and presented to the reader in a visual format. The authors of IRV-HC~\cite{Spytkowski2012} focused on highlighting how the additional properties of the proposed method impact the final result regarding statistical characteristics, not external validation. In the case of Hk++~\cite{Olech2016}, the authors attempted to find a way to verify the resulting hierarchies by generating synthetic hierarchical datasets and evaluating results by a modified F-Measure. However, the synthetically generated data used for testing was too simple to support a comprehensive evaluation.

\section{A brief introduction to Object Cluster Hierarchies}
\label{sec:intro_to_och}
Object Cluster Hierarchy~\cite{Spytkowski2012,Olech2016} is a recent variant of Hierarchical Clustering. The HC paradigm \cite{ABDOLALI2020333,doi:10.1137/1.9781611975031.26,Costa:2013,NIPS2017_6902,doi:10.1002/widm.53}, whether agglomerative or divisive, produces a dendrogram showing all levels of aggregations. Although there is a hierarchy relationship between nodes in a dendrogram, there is \textit{no hierarchy relationship between objects}. It is because all objects are assigned only to the dendrogram leaves, and clusters are generated by cutting the tree at any particular level. The level at which a dendrogram is cut determines the number of clusters in the final solution. Any node in the tree, except for leaves, does not have objects assigned to it. Thus, the structure of the generated clusters is flat. 

The OCH paradigm extends HC by allowing objects to be assigned to any node in the hierarchy tree. Researchers have already developed methods with such capability, enabling the hierarchy relation between clustered objects to be obtained, e.g.,~\cite{ghahramani2010tree,2010_bayesian_rose_trees,Spytkowski2012,Olech2016}.

    Within this paradigm, we have formulated three critical requirements~\cite{Spytkowski2012} to reflect a semantic (ontological) approach to Hierarchical Clustering:
\begin{enumerate}
    \item {\textit{Inheritance} -- every object that belongs to a given group also belongs to the parents' groups, up to the root;}
    \item {\textit{Retention} -- objects are not required to be located in the tree leaves;}
    \item {\textit{Variance} -- groups situated lower in the hierarchy are more specific, i.e., every child cannot have higher variation than its ancestors.}
\end{enumerate}

\noindent These requirements characterise human perception of hierarchy that can also be found in images~\cite{ILSVRC15,ghahramani2010tree}, documents~\cite{ghahramani2010tree,2010_bayesian_rose_trees}, and community structures in social networks~\cite{8508260, massaro2014hierarchical}.

\begin{figure}[ht!]
  \centering
  \begin{subfigure}[t]{0.5\textwidth}
  	\centering
    \includegraphics[width=0.70\textwidth]{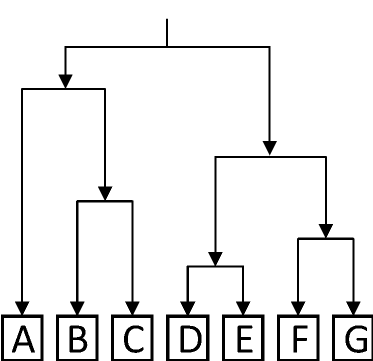}
    \caption{}
	%\caption{Hierarchical clustering example. The partial relation exists only between clusters.} 
	\label{fig:hierarchical-example} 
  \end{subfigure}%
  ~
  \begin{subfigure}[t]{0.5\textwidth}
  	\centering
    \includegraphics[width=0.40\textwidth, trim={0 1.15cm 0 1.15cm}, clip]{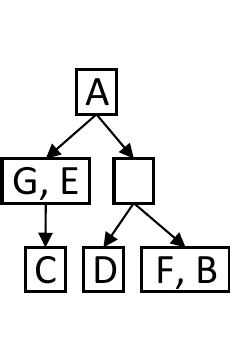}
    \caption{}
	%\caption{Object Cluster hierarchy example. The partial relation exists between both the clusters and the objects.}
	 \label{fig:och-example} 
  \end{subfigure}
  \caption{Examples of a dendrogram (a) and an Object Cluster Hierarchy (b). Letters represent objects and squares are groups. The arrows show the partial order relation. Note, that the diagrams represent \textit{the final location} of objects within the structures, but conceptually, every object belongs to all its predecessor groups (including the root). In the dendrogram, the partial order relation exists only between the clusters, whereas in the Object Cluster Hierarchy the partial order relation exists between both the clusters and the objects.}
  \label{fig:och-hierarchy-examples}
\end{figure}

In \Cref{fig:och-hierarchy-examples}, a comparison between Hierarchical Clustering (a) and Object Cluster Hierarchy (b) is presented based on a simple example. In the former, the final clustering is flat, and the number of clusters depends on the level where the dendrogram is cut. By cutting the tree from~\Cref{fig:hierarchical-example} at the bottom of the hierarchy, a set of seven clusters is formed, each of them containing one object. Regardless of where the hierarchy is cut, the resulting clustering consists of the same seven objects. In comparison, in \Cref{fig:och-example}, the whole OCH represents clustering --- partition of all seven letters. There is no need to cut such a hierarchy. Due to hierarchical relations, objects from child clusters conceptually belong to the parent clusters. Root always contains all the objects (i.e., the whole set), whereas leaves contain only what belongs to them.

\noindent

Hierarchical Clustering dendrogram always contains all the data points at the bottom of the hierarchy (in leaves). A hierarchy where data points are placed only in leaves is also a valid OCH. Thus, any possible HC result is always an OCH. However, the opposite does not always hold true due to the \textit{Retention} requirement in OCH. Hence, Object Cluster Hierarchy is an abstraction over Hierarchical Clustering.

In one of the early papers on the subject of OCH~\cite{ghahramani2010tree}, the authors pointed out that many data arise from a latent hierarchy, for example, a set of text documents or images. An OCH can model such data. In that paper the authors proposed a nonparametric method allowing to discover trees of unbounded width and depth by inferring them during a learning process. That method can assign objects to any node so some nodes can be empty, i.e., without any objects assigned. However, this method is not capable of generating OCH as it does not satisfy the \textit{Variance} requirement -- objects belonging to a child node can vary more than objects assigned to the ancestor nodes of that child. Starting from this method as the prototype, the ongoing development has been carried out to propose an improved version satisfying that requirement. The results were presented in~\cite{Spytkowski2012}.

The newly developed method couldn't be comprehensively compared with others against OCH-related characteristics. The problem, to the best of our knowledge, is twofold. Firstly, there are no appropriate evaluation measures dedicated to Object Cluster Hierarchies that would account for the distinguishing characteristics. Secondly, there are no available benchmark datasets with known properties, which could be used for testing different OCH generation methods. The former problem was partially addressed in~\cite{Spytkowski2016} where several external measures dedicated to OCH have been presented. Further, new external and internal validation techniques have been developed and are planned to be published soon, but their experimental studies require appropriate testing datasets. These two issues led to the development of a method generating synthetic data structures with known characteristics that are presented in this article. Ability to model the attributes of created datasets allows for more in-depth analysis of both methods and evaluation measures.
\section{Generator model}
\label{model}
    The data generated by this model can be interpreted as coming from an infinite mixture model. Most commonly an infinite mixture model is composed of infinite, indexed distributions from which the data is drawn. In such case the mixture weights can be drawn from a Dirichlet distribution, using the Stick Breaking Process. Generally, the distributions for the mixture components are unrelated to each other. This generator uses a similar approach. However, it bases its mixture weights on the Tree-Structured Stick Breaking Process, as described in~\cite{ghahramani2010tree}, which arranges the mixture components into a hierarchical structure where each component is a separate node. This structure also defines the relationship between the mixture components as children distributions are based on their parent's distribution parameters.
    
    Hierarchies generated using the Stick Breaking Process possesses the following characteristics~\cite{ghahramani2010tree, Spytkowski2012}:
\begin{enumerate}
    \item Every node can have an unlimited number of child nodes, thus the hierarchy depth and breadth are not limited;
    \item The children of a node are indexed and ordered. However, this indexing is not important after the generation process finishes;
    \item The hierarchy may contain empty nodes, that is, nodes that do not have any data directly assigned to them. However, such nodes might still have data belonging to them indirectly through the \textit{Inheritance} requirement;
    \item A child node distribution parameters are generated based on its parent distribution parameters and a kernel describing the transition;
    \item The shape of the generated hierarchy depends on several control hyperparameters described in~\Cref{parameter}.
\end{enumerate}
    
    Throughout this paper the following symbols are used to describe the generator and the generated model:
\begin{tabbing}
    $X$ \hspace{10mm} \=\;\;\; - set of all data points, or objects, \\
    $x_i$ \>\;\;\; - an object or data point with unique identifier $i$ represented by \\ \>\;\;\;\;\;\, a vector of features, \\
    $\Theta$ \>\;\;\; - set of all clusters, \\
    $\epsilon$ \>\;\;\; - specific cluster from $\Theta$, \\
    $\epsilon_{x_i}$ \>\;\;\; - cluster of object $x_i$, \\
    $\epsilon\epsilon_i$ \>\;\;\; - the $i$-th child of cluster $\epsilon$, if the Object Cluster Hierarchy is \\ \>\;\;\;\;\;\, defined,\\
    $\epsilon_\varnothing$ \>\;\;\; - the root cluster of the Object Cluster Hierarchy,\\
    $X_\epsilon$ \>\;\;\; - set of all objects in cluster $\epsilon$, \\
    $X_{E_\epsilon}$ \>\;\;\; - set of all objects in hierarchy subtree starting with node $\epsilon$, \\
    $|S|$ \>\;\;\; - number of elements in set \textit{S}, \\
    $|\epsilon|$ \>\;\;\; - a depth of a node $\epsilon$, \\
    $\theta_\epsilon$ \>\;\;\; - distribution of a node $\epsilon$. Specifically, $\theta_{\epsilon_\varnothing}$ is \\ \>\;\;\;\;\;\, a distribution of the root node, $\theta_{\epsilon_c}$ is a distribution of a node \\ \>\;\;\;\;\;\, $\epsilon_c$, and $\theta_{\epsilon\epsilon_i}$ is a distribution of an $i$-th child of cluster $\epsilon$, \\
    $Beta(\alpha, \beta)$ \; - Beta distribution with shape parameters $\alpha$ and $\beta$, \\
    $Gauss(\mu, \sigma)$ - Gaussian distribution with mean $\mu$ and standard deviation $\sigma$.
\end{tabbing}    
    Additionally, to the symbols above, several values are provided to the generator as parameters. The use of these parameters is further described in~\Cref{parameter}, and their influence on the final result is empirically shown in~\Cref{experiments}:
\begin{tabbing}
    $d$ \hspace{10mm} \=\;\;\;\; - the dimensionality of the generated data points, \\
    $n$ \>\;\;\;\; - the number of data points to be generated, \\
    $\alpha_0, \lambda$ \>\;\;\;\; - input parameters controlling the hierarchy depth, used by \\ \>\;\;\;\;\;\;\, equation $\alpha(\epsilon) = \alpha_0\lambda^{|\epsilon|}$, \\
    $\gamma$ \>\;\;\;\; - parameter controlling the width of a tree structure, \\
    $p, q$ \>\;\;\;\; - parameters controlling the specificity of the generated data; \\ \>\;\;\;\;\;\;\, they influence how much smaller the deviation of points' \\ \>\;\;\;\;\;\;\, features in the child node should be in comparison with points \\ \>\;\;\;\;\;\;\, in the parent node, \\
    $\theta_{\epsilon_\varnothing}$ \>\;\;\;\; - the distribution of the root node.
\end{tabbing}
    
    Two conditional probabilities are used to determine from which node data is generated. The first is the conditional probability of a datum remaining in node $\epsilon$, at depth $|\epsilon|$, when entering the node:
\begin{equation}
\nu_\epsilon = P(x \in X_\epsilon | x \in X_{E_\epsilon}),
\end{equation}
\begin{equation}\label{eq:alpha_function}
\nu_\epsilon \sim Beta(1, \alpha(\epsilon)),\;\;\; \alpha(\epsilon) = \alpha_0\lambda^{|\epsilon|}.
\end{equation}
    The second is the conditional probability of a datum being transferred to the subtree $\epsilon\epsilon_i$ if it does not remain in node $\epsilon$ and hasn't been transferred to any of the previous siblings (i.e., did not travel down sibling subtrees with a lower indices $\epsilon\epsilon_j,~j < i$): 

\begin{equation}
\begin{multlined}
\psi_{\epsilon\epsilon_i} = P(x \in X_{E_{\epsilon\epsilon_i}} | x \in X_{E_\epsilon} \wedge  x \not\in X_\epsilon \wedge \neg \exists_{j < i} x \in X_{E_{\epsilon\epsilon_j}}),
\end{multlined}
\end{equation}
\begin{equation}
\label{eq:beta_one_gamma}
\psi_{\epsilon\epsilon_i} \sim Beta(1, \gamma).
\end{equation}
    Additionally, we need to define the kernel. We begin with a specified root node distribution $\theta_{\epsilon_\varnothing}$ given as a starting parameter of the generation method. The values set at this point are the means $(\mu)$ and standard deviations $(\sigma)$ for each of the Gaussian distributions in the $d$ different dimensions:
\begin{equation}
\label{eq:theta-epsilon-varnothing}
\theta_{\epsilon_\varnothing} = (Gauss(\mu_{{\epsilon_\varnothing} 1}, \sigma_{{\epsilon_\varnothing} 1}), ... , Gauss(\mu_{{\epsilon_\varnothing} d}, \sigma_{{\epsilon_\varnothing} d})).
\end{equation}
    From there, for any node for which we need the distribution, we can draw the distribution based on the parent's distribution. The child's mean values are drawn directly from the parent distribution, and the child's standard deviation is based on a scaling factor ($\Delta\sigma_n$) drawn from the Beta distribution. The values are taken separately for each dimension:
    \begin{equation}
    \label{eq:theta-epsilon-epsilon-i}
    \begin{multlined}
    \theta_{\epsilon\epsilon_i} = (Gauss(\Delta\mu_1, \sigma_{\epsilon 1} \Delta\sigma_1), ... , Gauss(\Delta\mu_d, \sigma_{\epsilon d} \Delta\sigma_d)),
    \end{multlined}
    \end{equation}
    \begin{equation}
    \label{eq:delta-miu-n}
    \Delta\mu_n \sim Gauss(\mu_{\epsilon n}, \sigma_{\epsilon n}), \;\;\;n = 1, ..., d,
    \end{equation}
    
    \begin{equation}
    \label{eq:delta-sigma-n}
    \Delta\sigma_n \sim Beta(p, q), \;\;\;n = 1, ..., d.
    \end{equation}
    With the kernel defined we can now generate data from the model. We begin with the hyperparameters and the probability distribution for the root node ($\theta_{\epsilon_\varnothing} = (Gauss(\mu_{\epsilon_\varnothing 1}, \sigma_{\epsilon_\varnothing 1}), ..., Gauss(\mu_{\epsilon_\varnothing d}, \sigma_{\epsilon_\varnothing d}))$). \\
    
    \noindent The following process continues until $n$ points are generated:
    
    % \begin{addmargin}[1em]{0em}% 1em left, 0em right
    \begin{description}
        \item[Step 1:] If $|X| < n$ go to \textbf{Step2}, else \textbf{end}.
        \item[Step 2:] Randomly draw an insertion point $i_x \sim Uni(0, 1)$, $i_x \in (0, 1)$.
        \item[Step 3:] Set the root node as the current node ($\epsilon_c := \epsilon_\varnothing$), depth is 0 ($|\epsilon_c| := 0$).
        \item[Step 4:] If $\nu$ for the current node is not yet known, draw the value  $\nu_{\epsilon_c} \sim Beta(1, \alpha_0 \lambda ^ {|\epsilon_c|} )$.
        \item[Step 5:] If $i_x \leq \nu_{\epsilon_c}$ then $x \sim \theta_{\epsilon_c}$ ($x \sim \theta_{\epsilon_\varnothing}$ if $\epsilon_c = \epsilon_\varnothing$), the point belongs to the current node ($X_{\epsilon_c} := \{x\} \cup X_{\epsilon_c}$), go to \textbf{Step 1}, else move on to \textbf{Step 6}.
        \item[Step 6:] Adjust $i_x$ to new value: $i_x := (i_x - \nu_{\epsilon_c}) / (1 - \nu_{\epsilon_c})$.
        \item[Step 7:] Set the current child node index ($\epsilon_c\epsilon_i$) to the first child node of the current node: $i := 0$.
        \item[Step 8:] If $\psi$ for the current child node is not yet known, draw the value: $\psi_{\epsilon_c\epsilon_i} \sim Beta(1, \gamma)$.
        \item[Step 9:] If $\theta_{\epsilon_c\epsilon_i}$ for the current child node is not yet known, draw the values based on the parent of the node: \\
        $\theta_{\epsilon_c\epsilon_i} = (Gauss(\Delta\mu_1, \sigma_{\epsilon_c 1} \Delta\sigma_1), ..., Gauss(\Delta\mu_d, \sigma_{\epsilon_c d}\Delta\sigma_d))$,\\
        $\Delta\mu_1$ is drawn from the first dimension of the parent node ($\Delta\mu_1 \sim Gauss(\mu_{\epsilon_c 1}, \sigma_{\epsilon_c 1})$),\\
        \noindent        
        ... \\
        \noindent        
        $\Delta\mu_d$ is drawn from the $d$-th dimension of the parent node ($\Delta\mu_d \sim Gauss(\mu_{\epsilon_c d}, \sigma_{\epsilon_c d})$), \\
        \noindent        
        $\Delta\sigma_1\sim Beta(p, q)$, \\
        \noindent        
        ... \\
        \noindent        
        $\Delta\sigma_d\sim Beta(p, q)$.
        \item[Step 10:] If $i_x \leq \psi_{\epsilon_c\epsilon_i}$ go to \textbf{Step 11}, else go to \textbf{Step 12}.
        \item[Step 11:] Adjust the value of $i_x$ to new value: $i_x := i_x / \psi_{\epsilon_c\epsilon_i}$. Make the current child the current node ($\epsilon_c := \epsilon_c\epsilon_i$) and increase depth ($|\epsilon_c| := |\epsilon_c| + 1$). Go to \textbf{Step 4}.
        \item[Step 12:] Adjust the value of $i_x$ to new value $i_x := (i_x - \psi_{\epsilon_c\epsilon_i}) / (1 - \psi_{\epsilon_c\epsilon_i})$. Increment child index of currently relevant child node ($i: = i + 1$). Go to \textbf{Step 8}.
    \end{description}
    % \end{addmargin}
    
\begin{figure*}[h]
  \begin{center}
    \hspace*{-1.5cm}\includegraphics[clip, trim=5.25cm 12.8cm 5.4cm 4.4cm, width=1.15\textwidth]{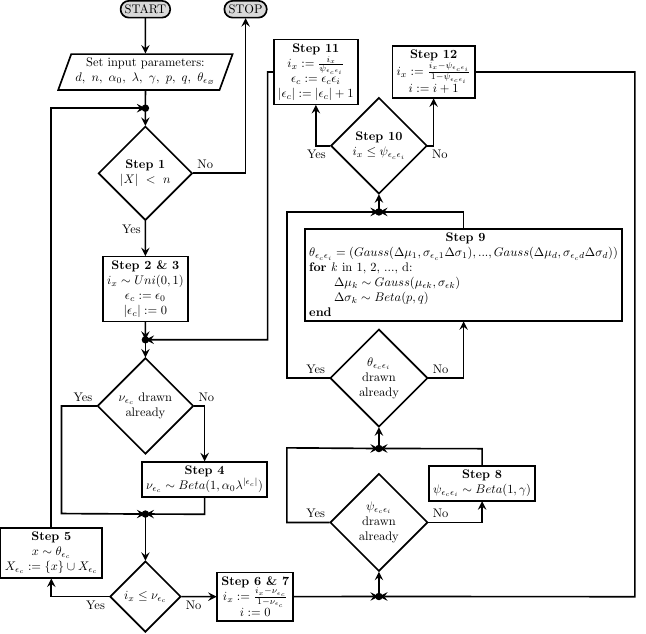}
	\caption{Block diagram of Object Cluster Hierarchy generator.} \label{fig:generator-block-diagram} 
  \end{center}
\end{figure*}

    \noindent The generation process described above is illustrated as a block diagram in Figure~\ref{fig:generator-block-diagram}.
    \noindent
    \clearpage
    
	\section{Parameter selection}
    \label{parameter}    
    Hierarchical data might follow hierarchies of different characteristics, e.g., depth, width, the average number of objects per node. Thus, to be applicable to various problems, the generator should provide high flexibility in generating a variety of hierarchies. The primary interest is in the structure of the hierarchy, that is whether the hierarchy is tall or short, wide or narrow, as well as the distribution of data across the levels of the hierarchy. Additionally, the difference between data in parent and child nodes can also be important in some cases. All of these are controlled by several parameters in the model:
    \begin{enumerate}
        \item hierarchy depth: $\alpha_0$, $\lambda$ or in a more general sense, the $\alpha(\epsilon)$ function,
        \item hierarchy width: $\gamma$,
        \item data specificity: $p$, $q$,
        \item root node distribution: $\theta_{\epsilon_\varnothing}$.
    \end{enumerate}
    The remaining part of this section is organised into five subsections. In the following four sub-sections all the above-mentioned parameter groups are discussed. In~\Cref{sec:controlling-the-hierarchy-depth} the $\alpha(\epsilon)$ function and its influence on the depth of hierarchy are presented. The $\gamma$ parameter and its impact on the hierarchy width are described in~\Cref{sec:controlling-hierarchy-width}. The differences in data distributions between nodes in a hierarchy (especially parent-child nodes) are discussed in~\Cref{sec:controlling_data_specificity}. As the hierarchy generation process is iterative and top-down, the initial (root) distribution parameters have an impact on the final hierarchy -- this is discussed in~\Cref{sec:influence_of_starting_distribution}. A reassignment post-processing procedure allowing for the generated hierarchies to be denoised is described in~\Cref{sec:reassignment_post_processing}. Furthermore, the influence of parameters on the generated trees discussed theoretically in this section is also empirically demonstrated later in the paper. 
    \subsection{Controlling hierarchy depth}
    \label{sec:controlling-the-hierarchy-depth}
    The depth of the hierarchy is controlled by the function $\alpha(\epsilon) = \alpha_0\lambda^{|\epsilon|}$. The higher the probability of a datum remaining in a node, the fewer data will travel deep down the tree, and thus the tree will be shallower. On the other hand, if the probability is low, the data will, on average, travel deeper into the tree. The average probability of data remaining in a given node is based on the selected $\alpha$ function, which influences the structure of the tree based on the $\alpha_0$ and $\lambda$ parameters:
    \begin{equation}
    \mathbf{E}[x \in X_\epsilon, |\epsilon| = 0] = \frac{1}{1 + \alpha_0},
    \end{equation}
    \begin{equation}
    \mathbf{E}[x \in X_\epsilon, |\epsilon| = n] = \frac{\prod_{i=0}^{n-1} \alpha_0\lambda^i}{\prod_{j=0}^{n}(1 + \alpha_0\lambda^j)}.
    \end{equation}
    Additionally the variance can also be calculated:
    \begin{equation}
    \mathbf{var}[x \in X_\epsilon, |\epsilon| = 0] = \frac{\alpha_0}{(1 + \alpha_0)^2(2 + \alpha_0)},
    \end{equation}
    \begin{equation}
    \begin{multlined}
    \mathbf{var}[x \in X_\epsilon, |\epsilon| = n] = \\ = \frac{2 \prod_{i = 0}^{n-1}\alpha_0\lambda^i}{(1+\alpha_0\lambda^n)\prod_{j=0}^{n}(2+\alpha_0\lambda^j)} \\ - \left(\textbf{E}[x \in X_\epsilon, |\epsilon| = n]\right)^2.
    \end{multlined}
    \end{equation}
    It is possible to predict the shape of the tree and data distribution based on $\alpha_0$ and $\lambda$ parameters:
    \begin{itemize}
        \item $\alpha_0 = 1, \lambda = 1$: the structure of the tree is chaotic and hard to predict, the further away the parameters move from these values the more stable the tree becomes;
        \item $\alpha_0 < 1, \lambda \leq 1$: shallow structure, data located primarily at the top of the tree;
        \item $\alpha_0 \leq 1, \lambda > 1$: similar to the above case, the depth of the tree increases but most data is located at the top of the tree;
        \item $\alpha_0 \geq 1, \lambda < 1$: the structure is deep, but data is not located at the top, the bigger $\alpha_0$ starts out, and smaller $\lambda$ is the more data will move down the tree into the central or lower region;
        \item $\alpha_0 > 1, \lambda \geq 1$: deep structure, data located primarily at the top of the tree but spread out.
    \end{itemize}
    \subsection{Controlling hierarchy width}
    \label{sec:controlling-hierarchy-width}
    The width of the tree is based on the value of the $\gamma$ parameter at a given node depth $j$. Given that 
    $x \in X_{E_\epsilon}$ and $x \not \in X_\epsilon$ the average probability of data being generated from a specific subtree (based on the index) can be used to estimate the number of children a node can potentially have:
    \begin{equation}
    \mathbf{E}[x \in X_{E_{\epsilon\epsilon_i}}, i = 1] = \frac{1}{1 + \gamma},
    \end{equation}
    
    \begin{equation}
    \mathbf{E}[x \in X_{E_{\epsilon\epsilon_i}}, i = n] = \frac{\gamma^{n-1}}{(1 + \alpha_0\lambda^j)^n}.
    \end{equation}
    Variance for these values can also be calculated:
    \begin{equation}
    \mathbf{var}[x \in X_{E_{\epsilon\epsilon_i}}, i = 1] = \frac{\gamma}{(1 + \gamma)^2(2 + \gamma)},
    \end{equation}
    
    \begin{equation}
    \begin{multlined}
    \mathbf{var}[x \in X_{E_{\epsilon\epsilon_i}}, i = n] = \frac{2 \gamma^{n-1}}{(1+\gamma)(2+\gamma)^n} - \left(\textbf{E}[x \in X_{E_{\epsilon\epsilon_i}}, i = n]\right)^2.
    \end{multlined}
    \end{equation}
    Influence of the parameter $\gamma$ on the generated hierarchies is as follows:
    \begin{itemize}
        \item $\gamma = 1$: the number of children is chaotic and difficult to predict;
        \item $\gamma < 1$: narrower tree, fewer children per node on average;
        \item $\gamma > 1$: wider tree, more children per node on average.
    \end{itemize}
    \subsection{Controlling data specificity}
    \label{sec:controlling_data_specificity}
    When a new group is considered, the parameters for that group data points distribution are drawn based on the parent distribution and the kernel parameters $p$ and $q$. An important aspect of the generated model is that data becomes more specific at lower nodes following the OCH principles. However, the values that are taken for the kernel change the average proportion 
    %between data standard deviation between the parent and child. 
    of standard data deviation between the parent and child.
    This is based on the expected standard deviation of the new node compared to the old node (taken separately in each dimension):
    \begin{equation}\label{eq:analitical_expected_standard_dev}
    \mathbf{E}[\sigma_{\epsilon\epsilon_id}] = \sigma_{\epsilon d} \frac{p}{p + q},
    \end{equation}
    \begin{equation}\label{eq:analitical_variance_standard_dev}
    \mathbf{var}[\sigma_{\epsilon\epsilon_id}] = \sigma_{\epsilon d} \frac{pq}{(p + q)^2(p + q + 1)}.
    \end{equation}
    By selecting $p$ and $q$, the rate at which the nodes become more specific can be altered. The lower the mean is, the more specific every child will be (on average), the higher the variance is, the more variety there will be in how the child nodes relate to their parent.
    \subsection{Influence of starting distribution on results}
    \label{sec:influence_of_starting_distribution}
    Due to the relative nature of the model (i.e., the specific values generated from the model are calculated relative to each other, starting from the root distribution, as shown in equations~\ref{eq:theta-epsilon-varnothing},~\ref{eq:theta-epsilon-epsilon-i},~\ref{eq:delta-miu-n},~\ref{eq:delta-sigma-n}), the choice of initial distribution parameters is not very important. The data generated from the model can be scaled afterwards to any desired values as well as moved in any direction along any dimension. Because of this, the generator assumes a following data distribution for the root node:
    \begin{equation}
    \theta_{\epsilon_\varnothing} = (Gauss(\mu_{{\epsilon_\varnothing} 1}, \sigma_{{\epsilon_\varnothing} 1}), ... , Gauss(\mu_{{\epsilon_\varnothing} d}, \sigma_{{\epsilon_\varnothing} d})),
    \end{equation}
    where
    \begin{equation}
         \mu_{{\epsilon_\varnothing} 1} = \mu_{{\epsilon_\varnothing} 2} = ... = \mu_{{\epsilon_\varnothing} d} = 0,
    \end{equation}
    and
    \begin{equation}
         \sigma_{{\epsilon_\varnothing} 1} = \sigma_{{\epsilon_\varnothing} 2} = ... = \sigma_{{\epsilon_\varnothing} d} = \sigma_{max}.
    \end{equation}
    Every dimension of the root node is described by a normal distribution with zero mean and the standard deviation of value $\sigma_{max}$ which is a method's parameter. Data generated from the model can be then post-processed to a more desirable spread of values. This is done by applying scaling and translation to all the data generated by the model as well as the parameters of each group node.
    \subsection{Reassignment post-processing}
    \label{sec:reassignment_post_processing}
    As it is shown in~\Cref{model}, the assignment of points to groups is conducted in a top-down manner separately for every point. It starts from the root node and moves down the hierarchy a particular path considering a different sequence of nodes for a point assignment. For every visited node the conditional probabilities $\nu_\epsilon$ and $\psi_{\epsilon\epsilon_i}$ are calculated (\Cref{eq:theta-epsilon-epsilon-i,eq:beta_one_gamma}) and their values determine which sequence of nodes is considered before a point is finally assigned. A point is assigned to the first node for which certain conditions are met (see \textbf{Step 5} in~\Cref{model}). As this process is greedy and stochastic, only a subset of nodes will be considered for point assignment which does not guarantee that a node with the highest probability will be selected. It introduces noise to the nodes and biases their data distributions.
    
    To address such behaviour, a hierarchy can undergo one form of post-processing after being generated. This process, referred in this paper as \textit{reassignment}, moves the data between clusters in such a way that each object belongs to the cluster it is most likely to be generated from:
    \begin{equation}
    \forall_{x \in X}\left( x \in X_{\epsilon_a}\Leftrightarrow \neg\exists_{\epsilon_b \neq \epsilon_a}L(x|\theta_{\epsilon_a})<L(x|\theta_{\epsilon_b})\right )
    \end{equation}
    The process does not modify the number of clusters, hierarchy relations between them or their parameters in any way. It merely relocates data to reduce noise and produce cleaner clusters.
    
	\section{Experiments}
	\label{experiments}
	The generator was tested with a number of different goals in mind. The tests serve to empirically investigate the analytical and intuitive properties of the introduced parameters provided in the previous sections. Thus, a large part of the experiments serves to verify how the different parameter values affect the generated hierarchies. Further, the experiments aim at demonstrating various properties of the generated hierarchies and generator flexibility in producing differently-structured hierarchical data. The influence of the reassignment post-processing (\Cref{sec:reassignment_post_processing}) on the generated hierarchies has also been investigated. The goal of the reassignment process is to reduce noise in the generated data by moving objects to the node for which the likelihood of being drawn from is the highest. A comparison of post-processed hierarchies with unmodified ones was performed. All of these experiments were done with a primary objective to provide a potential user with a comprehensive understanding of the generator and to assist him/her in choosing the best parameter set for any given use case. An additional benefit from conducted experiments is the establishment of new benchmarking datasets for OCH that are ready to use for method comparison without a need to use the generator.
    
    Due to the stochastic nature of the generator, the results presented in this section were obtained by averaging over 100 generated hierarchies for each of the used parameter sets (\textit{s00} -- \textit{s07}) shown in~\Cref{tab:parameter-configurations}. Each of the used parameter sets represents a different type of hierarchical structure. Some parameters such as $n$ = 10,000, $d = 2$, $p = 1$, $q = 5$, $\sigma_{min} = 0.05$ and $\sigma_{max} = 10$ remained constant across all experiments whereas $\alpha_0$, $\lambda$, $\gamma$ varied. For every generator run two hierarchies were produced: as generated from the statistical model (the initial assignment of data to nodes), and after reassignment of data (\textit{reassigned} datasets). Datasets with the initial assignment of data are referred to as \textit{s00} -- \textit{s07} depending on which parameter set-up was used (\Cref{tab:parameter-configurations}). For the reassigned hierarchies, the naming convention is the same with an additional letter \textit{'r'} (\textit{s00r} -- \textit{s07r}). Regardless of which variant of the hierarchy (with or without the reassigned procedure) has been created, generator parameters remain the same as the reassignment procedure is applied \textit{after} the data is created. In other words, the reassignment procedure does not influence the data generation process.
    \begin{table}[b!]
\setlength{\tabcolsep}{8pt}
\begin{center}
\caption{Generator parameters used to create experimental data sets from \textit{s00} to \textit{s07} (or  \textit{s00r} to \textit{s07r} correspondingly if the reassigned post-processing procedure has been used; the parameters are the same since reassignment is applied after the dataset is created). The remaining parameters are shared between test sets: $n$ = 10,000, $d = 2$, $p = 1$, $q = 5$, $\sigma_{min} = 0.05$ and $\sigma_{max} = 10$. Parameter set selection is based on previous research in this area~\cite{ghahramani2010tree}}.\label{tab:parameter-configurations}
\begin{tabular}{cccccccccc}
$Set$ & $\alpha_0$ & $\lambda$ & $\gamma$ \\
\hline
\textit{s00} and \textit{s00r} & 1 & 0.5 & 0.2 \\
\textit{s01} and \textit{s01r} & 1 & 1.0 & 0.2 \\
\textit{s02} and \textit{s02r} & 1 & 1.0 & 1.0 \\
\textit{s03} and \textit{s03r} & 5 & 0.5 & 0.2 \\
\textit{s04} and \textit{s04r} & 5 & 1.0 & 0.2 \\
\textit{s05} and \textit{s05r} & 5 & 0.5 & 1.0 \\
\textit{s06} and \textit{s06r} & 25 & 0.5 & 0.2 \\
\textit{s07} and \textit{s07r} & 25 & 0.5 & 1.0 \\
\end{tabular}
\end{center}
\end{table} %ta tabelka musi byc PRZED ponizzszymi, bo inaczej nie bedzie numerowana jako 1, a to troche dziwne, bo w tekscie sie najpierw do niej odwolujemy
    
    Several quantitative measures were used to investigate the aggregate properties of the generated hierarchies:
    \begin{itemize}
        \item $\bar{N}$ -- the number of nodes in the hierarchy, averaged over all generated hierarchies,
        \item $\bar{L}$ -- the number of leaves in the hierarchy (nodes with no children or with empty children only), averaged over all generated hierarchies,
        \item $\bar{D}$ -- the depth of the hierarchy, averaged over all generated hierarchies,
        \item $\bar{B}$ -- the breadth of the hierarchy, averaged over all levels in a hierarchy, and over all hierarchies generated,
        \item $\bar{P}$ -- the average length of all paths in a hierarchy, averaged over all generated hierarchies.
    \end{itemize}
    All of the reported average measures are accompanied by standard deviations. Since $B$ and $P$ are averages of averages instead of a standard deviation, an \textit{average} of standard deviations over all generated hierarchies is provided. All defined measures are reported separately for the initially generated (\Cref{tab:quantitative-hierarchy-analysis}) and reassigned hierarchies (\Cref{tab:quantitative-hierarchy-analysis-R}). The remaining experiments results (\Cref{fig:level-by-level-number-of-instances,fig:level-by-level-number-of-instances-R,fig:level-by-level-hierarchy-width,fig:level-by-level-hierarchy-width-R,fig:level-by-level-number-of-children-per-node,fig:level-by-level-number-of-children-per-node-R,fig:level-by-level-number-of-leaves,fig:level-by-level-number-of-leaves-R,fig:branching-factor-histogram,fig:branching-factor-histogram-R}) are presented as histograms averaged over the 100 generated hierarchies for each parameter set. The histograms (except for~\Cref{fig:branching-factor-histogram,fig:branching-factor-histogram-R}) present measures across different levels of hierarchies providing a more in-depth (structural) view on the generated data. \Cref{fig:level-by-level-number-of-instances,fig:level-by-level-hierarchy-width,fig:level-by-level-number-of-children-per-node,fig:level-by-level-number-of-leaves,fig:branching-factor-histogram} show results when the reassignment post-processing has not been executed, whereas \Cref{fig:level-by-level-number-of-instances-R,fig:level-by-level-hierarchy-width-R,fig:level-by-level-number-of-children-per-node-R,fig:level-by-level-number-of-leaves-R,fig:branching-factor-histogram-R} show values after the post-processing. The reported measures are:
    \begin{itemize}
        \item average width per level~(\Cref{fig:level-by-level-hierarchy-width,fig:level-by-level-hierarchy-width-R}), %2
        \item average number of objects per node per level~(\Cref{fig:level-by-level-number-of-instances,fig:level-by-level-number-of-instances-R}), %1
        \item average number of children per node per level~(\Cref{fig:level-by-level-number-of-children-per-node,fig:level-by-level-number-of-children-per-node-R}), %4
        \item average number of leaves per level~(\Cref{fig:level-by-level-number-of-leaves,fig:level-by-level-number-of-leaves-R}), %3
        \item average number of nodes with a given number of children~(\Cref{fig:branching-factor-histogram,fig:branching-factor-histogram-R}). %5
    \end{itemize}
    \begin{table*}[h!]
\setlength{\tabcolsep}{3pt}
\begin{center}
\caption{Accumulative characteristics of generated hierarchies \textbf{without} the reassignment procedure. Average $\bar{X}$ values together with standard deviation $\sigma_{\bar{X}}$ (or an average of standard deviations $\bar{\sigma}_{\bar{X}}$) are provided.\label{tab:quantitative-hierarchy-analysis}}
\begin{tabular}{crrrrrrrrrr}
\multirow{2}{*}{\textit{\small{Set}}} & \multicolumn{2}{c}{\textit{\small{Nodes}}} & \multicolumn{2}{c}{\textit{\small{Leaves}}} & \multicolumn{2}{c}{\textit{\small{Depth}}} & \multicolumn{2}{c}{\textit{\small{Breadth}}} & \multicolumn{2}{c}{\textit{\small{Path length}}} \\
& \multicolumn{1}{c}{$\bar{N}$} & \multicolumn{1}{c}{$\sigma_{\bar{N}}$} & \multicolumn{1}{c}{$\bar{L}$} & \multicolumn{1}{c}{$\sigma_{\bar{L}}$} & \multicolumn{1}{c}{$\bar{D}$} & \multicolumn{1}{c}{$\sigma_{\bar{D}}$} & \multicolumn{1}{c}{$\bar{B}$} & \multicolumn{1}{c}{$\bar{\sigma}_{\bar{B}}$} & \multicolumn{1}{c}{$\bar{P}$} & \multicolumn{1}{c}{$\bar{\sigma}_{\bar{P}}$} \\ 
\hline
\textit{s00} & 17.58 & 7.80 & 8.75 & 4.33 & 4.06 & 0.82 & 3.40 & 2.14 & 2.86 & 0.93 \\
\textit{s01} & 95.23 & 53.06 & 31.00 & 17.85 & 11.73 & 2.80 & 7.14 & 5.10 & 6.30 & 2.50 \\
\textit{s02} & 556.81 & 329.74 & 271.80 & 188.79 & 12.33 & 2.20 & 40.44 & 41.33 & 5.40 & 1.99 \\
\textit{s03} & 58.19 & 21.33 & 25.84 & 10.62 & 6.41 & 0.73 & 7.84 & 5.97 & 4.36 & 1.13 \\
\textit{s04} & 3090.88 & 944.13 & 483.62 & 187.81 & 52.54 & 6.83 & 58.14 & 59.92 & 19.39 & 7.88 \\
\textit{s05} & 485.43 & 149.83 & 297.62 & 108.54 & 6.88 & 0.55 & 61.69 & 64.09 & 4.25 & 1.00 \\
\textit{s06} & 175.71 & 61.50 & 67.57 & 26.01 & 8.83 & 0.64 & 17.84 & 14.97 & 6.20 & 1.32 \\
\textit{s07} & 2071.07 & 536.50 & 1109.17 & 367.70 & 9.27 & 0.51 & 201.88 & 223.79 & 5.88 & 1.16 \\
\end{tabular}
\end{center}
\end{table*}
\begin{table*}[h!]
\setlength{\tabcolsep}{3pt}
\begin{center}
\caption{Accumulative characteristics of generated hierarchies \textbf{with} the reassignment procedure. Average $\bar{X}$ values together with standard deviation $\sigma_{\bar{X}}$ (or an average of standard deviations $\bar{\sigma}_{\bar{X}}$) are provided.\label{tab:quantitative-hierarchy-analysis-R}}
\begin{tabular}{crrrrrrrrrr}
\multirow{2}{*}{\textit{\small{Set}}} & \multicolumn{2}{c}{\textit{\small{Nodes}}} & \multicolumn{2}{c}{\textit{\small{Leaves}}} & \multicolumn{2}{c}{\textit{\small{Depth}}} & \multicolumn{2}{c}{\textit{\small{Breadth}}} & \multicolumn{2}{c}{\textit{\small{Path length}}} \\
& \multicolumn{1}{c}{$\bar{N}$} & \multicolumn{1}{c}{$\sigma_{\bar{N}}$} & \multicolumn{1}{c}{$\bar{L}$} & \multicolumn{1}{c}{$\sigma_{\bar{L}}$} & \multicolumn{1}{c}{$\bar{D}$} & \multicolumn{1}{c}{$\sigma_{\bar{D}}$} & \multicolumn{1}{c}{$\bar{B}$} & \multicolumn{1}{c}{$\bar{\sigma}_{\bar{B}}$} & \multicolumn{1}{c}{$\bar{P}$} & \multicolumn{1}{c}{$\bar{\sigma}_{\bar{P}}$} \\ 
\hline
\textit{s00r} & 18.11 & 8.17 & 9.30 & 4.68 & 4.05 & 0.82 & 3.50 & 2.23 & 2.86 & 0.92 \\
\textit{s01r} & 98.56 & 54.99 & 34.76 & 19.50 & 11.70 & 2.82 & 7.40 & 5.31 & 6.30 & 2.48 \\
\textit{s02r} & 642.02 & 367.21 & 363.20 & 211.42 & 12.23 & 2.19 & 46.88 & 47.62 & 5.43 & 1.96 \\
\textit{s03r} & 59.88 & 22.11 & 27.64 & 11.33 & 6.40 & 0.74 & 8.07 & 6.15 & 4.35 & 1.14 \\
\textit{s04r} & 3099.03 & 936.39 & 597.36 & 176.93 & 52.05 & 6.96 & 58.90 & 60.44 & 19.17 & 7.78 \\
\textit{s05r} & 552.61 & 151.38 & 366.31 & 100.87 & 6.87 & 0.56 & 70.30 & 72.92 & 4.25 & 1.03 \\
\textit{s06r} & 180.47 & 63.97 & 73.14 & 28.05 & 8.81 & 0.65 & 18.34 & 15.40 & 6.16 & 1.35 \\
\textit{s07r} & 2310.21 & 524.57 & 1375.60 & 302.29 & 9.27 & 0.51 & 225.19 & 250.42 & 5.87 & 1.18 \\
\end{tabular}
\end{center}
\end{table*}
    
    The results of the conducted experiments can be confronted with prior analytical estimations of the effect that parameters have on the structure of the hierarchy (\Cref{parameter,sec:controlling-the-hierarchy-depth,sec:controlling-hierarchy-width,sec:controlling_data_specificity}). The simplest case is the $\gamma$ parameter (\Cref{sec:controlling-hierarchy-width}). This parameter is responsible for the formation of child nodes and as such, the breadth of the hierarchy. For datasets that differ only by the $\gamma$ value (\textit{s01} and \textit{s02} or \textit{s06} and \textit{s07}), the distribution of data per level is very similar~(\Cref{fig:level-by-level-hierarchy-width,fig:level-by-level-hierarchy-width-R}). It is because the data distribution is controlled by the $\alpha$ function, which is not influenced by $\gamma$. On the other hand, there is a significant change in the width of the hierarchy, approximately by one order of magnitude (10 times higher for higher $\gamma$), as it was predicted by the prior analysis.
    
    The influence of the $\alpha_0$ and $\lambda$ parameters on generated hierarchies is difficult to describe (\Cref{sec:controlling-the-hierarchy-depth}) as the two parameters are interwoven together within the $\alpha$ function~(\Cref{eq:alpha_function}); it also depends on the level of a hierarchy that is currently considered. However, the impact of this function is the best presented in~\Cref{fig:level-by-level-number-of-instances,fig:level-by-level-number-of-instances-R}, especially when results for datasets \textit{s00}, \textit{s01}, \textit{s02} and \textit{s04} are compared with the results for \textit{s03}, \textit{s05}, \textit{s06} and \textit{s07}. The first set has a clear tendency to retain data in higher nodes (nodes that are located closer to the root). In comparison, the other set has the main mass of data located in the lower nodes (nodes located distant from the root). Especially with the \textit{s06} and \textit{s07} the majority of objects are located in lower nodes, close to the 5th level. For these two datasets, we can see that $\alpha$ on average starts out with low values (due to a high value of $\alpha_0$) and because of $\lambda$ parameter being smaller than 1, inclines as moving lower in the hierarchy (compare the influence of parameters on the Beta distribution\footnote{\url{http://eurekastatistics.com/beta-distribution-pdf-grapher/}}). For a low value of $\alpha$, the probability of retaining data in a node is on average low (see steps 2-5 in~\Cref{fig:generator-block-diagram}). Thus the nodes close to the root do not retain data, but as the value of $\alpha$ increases, more objects gather in the lower nodes of the hierarchy before eventually, the remaining data is passed on to the lowest nodes (leaves). From this, we come to an important conclusion about the importance of these two parameters. In cases where it is undesirable to have many generic (root level) objects, and it is important to have clearly distinct, specific (lower level) objects parameters $\alpha_0$ and $\lambda$ must have values similar to those present in \textit{s06} and \textit{s07} -- high $\alpha_0$ and $\lambda$ that controls the decline of the \textit{Beta} function value over levels to be smaller than 1. Such behaviour was earlier predicted from the analytical study of the parameters~(\Cref{sec:controlling-the-hierarchy-depth}), and the conducted experiments show that behaviour empirically. It appears that the bulk of data is retained at the level in which $\alpha(\epsilon)$ drops below $1$.
    
\begin{figure}[H]
        \centering
        \includegraphics [scale=1.5] {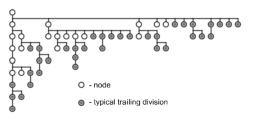}
        \caption{A simple schema of the location of \textit{trailing divisions}.}
        \label{fig:trailing}
    \end{figure}

    A very prominent behaviour of the generator seen in all test cases is the production of -- what will be referred to from this point onward -- \textit{trailing divisions} of data. Trailing divisions occur when the generator attempts to split small remaining partitions of data. This happens both in the right (higher index) children of a populated node and lowers down the hierarchy as presented in~\Cref{fig:trailing}. In both cases, it is possible to observe large numbers of nodes with a low number of children, usually one or zero (in the latter case the node becoming a leaf node), as well as many nodes that are not populated with data. Trailing divisions reveal the fractal nature of trees generated by the procedure, which manifests itself both when producing direct children for a node (horizontal self-similarity) and going down the hierarchy (vertical self-similarity).
    The above phenomenon can be visualised horizontally as an ordered set of all children of any node being statistically similar to the ordered set of all children of the node except the first one. It is a direct effect of the Tree-Structured Stick Breaking process (TSSB)~\cite{ghahramani2010tree}. Similarly to the above, from a vertical point of view, in any tree with $\lambda = 1$ (the conditional probability of a datum being assigned to a particular node is constant for all the nodes, see~\Cref{eq:alpha_function}) all sub-trees of a node are statistically similar to that node. In the presented experimental data, these trailing divisions are visible as the falling off "tail" in the histograms of data instances distribution per level  (\Cref{fig:level-by-level-number-of-instances,fig:level-by-level-number-of-instances-R}) as well as the cause of the high deviation of width within the trees (\Cref{fig:level-by-level-hierarchy-width,fig:level-by-level-hierarchy-width-R}). Unfortunately, due to the nature of the TSSB distribution, it is impossible to avoid this behaviour without post-processing. No forms of such post-processing were employed for the experiments presented in this paper.
    
    Since the $\gamma$ value does not change between levels (see \Cref{eq:beta_one_gamma}), a critical factor in considering how many children a node will have is the number of data passing through the node during the hierarchy generation process. Because of that, nodes that are located closer to the root are more likely to have more children than smaller nodes lower down the hierarchy. It has also been verified experimentally, and it is shown in~\Cref{fig:level-by-level-number-of-children-per-node,fig:level-by-level-number-of-children-per-node-R}. Additionally, the more children a node has, the more of them will be small nodes, i.e., nodes through which few objects pass, resulting primarily in leaf nodes or nodes with a single child. This behaviour of smaller nodes also transfers lower down the hierarchy where less data reaches, leading to similar behaviour.
    
    Finally, the reassigned test cases show a tendency for data to move down hierarchy levels. It is best shown when comparing \Cref{fig:level-by-level-number-of-instances} with \Cref{fig:level-by-level-number-of-instances-R}. Intuitively, the groups located lower down in the hierarchy are more specific and potentially conflicting data would be prone to moving down into the more specific child clusters during the reassignment process. However, despite this tendency, the hierarchies structure do not change. Hierarchies retain most of their initial characteristics with the mass of data being shifted down towards the lower levels. Due to the post-processing applied to these datasets, they can be better suited for initial testing of grouping methods as they contain less noise. Testing using both types of datasets (unfiltered and filtered) may be the preferred and most valuable approach in every case where the features of the objects are considered. 
    %%%%%%%%%%%%%%%%%%%%%%%%%%%%%%%%%%%%%%%%%%%%%%%%%%%%%%%%%%%%%%%%%%%%%%%%%%%%%%%%%%%%%%%%%%%%%%%%%%%%%%%%%%%%%%
    \noindent
\begin{figure*}[p!]
\centering
\begin{tabular}{rrrr}
%%%%%%%%%%%%%%%%%%%%%%%%%%%%%%%%%%%%%%%%%%%%%%%%%%%%%
  \begin{tikzpicture}
    \begin{axis} 
		[
	  	ybar,
	  	title=s00,
		  xmin = -0.5,
		  xmax = 6.5,
		  bar width = 11.6,
		]
		\addplot
		[
			color = blue,
			fill = blue,
		]
		table
		[
	  	col sep = semicolon,
	  	x = "Level",
	  	y = "s00",
		]
		{avgHierarchyWidthPerLevel.csv};
		\end{axis}
  \end{tikzpicture}&%
%%%%%%%%%%%%%%%%%%%%%%%%%%%%%%%%%%%%%%%%%%%%%%%%%%%%%
  \begin{tikzpicture}
    \begin{axis} 
		[
	  	ybar,
	  	title=s01,
		  xmin = -0.6,
		  bar width = 4.2,
		]
		\addplot
		[
			color = blue,
			fill = blue,
		]
		table
		[
	  	col sep = semicolon,
	  	x = "Level",
	  	y = "s01",
		]
		{avgHierarchyWidthPerLevel.csv};
		\end{axis}
  \end{tikzpicture}&%
%%%%%%%%%%%%%%%%%%%%%%%%%%%%%%%%%%%%%%%%%%%%%%%%%%%%%
  \begin{tikzpicture}
    \begin{axis} 
		[
	  	ybar,
	  	title=s02,
		  xmin = -0.55,
		  bar width = 4.4,
		]
		\addplot
		[
			color = blue,
			fill = blue,
		]
		table
		[
	  	col sep = semicolon,
	  	x = "Level",
	  	y = "s02",
		]
		{avgHierarchyWidthPerLevel.csv};
		\end{axis}
  \end{tikzpicture}&%
%%%%%%%%%%%%%%%%%%%%%%%%%%%%%%%%%%%%%%%%%%%%%%%%%%%%%
  \begin{tikzpicture}
    \begin{axis} 
		[
	  	ybar,
	  	title=s03,
		  xmin = -0.5,
		  xmax = 8.5,
	   	bar width = 9.0,
		]
		\addplot
		[
			color = blue,
			fill = blue,
		]
		table
		[
	  	col sep = semicolon,
	  	x = "Level",
	  	y = "s03",
		]
		{avgHierarchyWidthPerLevel.csv};
		\end{axis}
  \end{tikzpicture}\\%
%%%%%%%%%%%%%%%%%%%%%%%%%%%%%%%%%%%%%%%%%%%%%%%%%%%%%
  \begin{tikzpicture}
    \begin{axis} 
		[
	  	ybar,
	  	title=s04,
		  xmin = -0.5,
		  bar width = 0.8,
		]
		\addplot
		[
			color = blue,
			fill = blue,
		]
		table
		[
	  	col sep = semicolon,
	  	x = "Level",
	  	y = "s04",
		]
		{avgHierarchyWidthPerLevel.csv};
		\end{axis}
  \end{tikzpicture}&%
%%%%%%%%%%%%%%%%%%%%%%%%%%%%%%%%%%%%%%%%%%%%%%%%%%%%%
  \begin{tikzpicture}
    \begin{axis} 
		[
	  	ybar,
	  	title=s05,
		  xmin = -0.45,
		  xmax = 9.3,
		  bar width = 8.1,
		]
		\addplot
		[
			color = blue,
			fill = blue,
		]
		table
		[
	  	col sep = semicolon,
	  	x = "Level",
	  	y = "s05",
		]
		{avgHierarchyWidthPerLevel.csv};
		\end{axis}
  \end{tikzpicture}&%
%%%%%%%%%%%%%%%%%%%%%%%%%%%%%%%%%%%%%%%%%%%%%%%%%%%%%
  \begin{tikzpicture}
    \begin{axis} 
		[
	  	ybar,
	  	title=s06,
		  xmin = -0.5,
		  xmax = 10.6,
		  bar width = 7.1,
		]
		\addplot
		[
			color = blue,
			fill = blue,
		]
		table
		[
	  	col sep = semicolon,
	  	x = "Level",
	  	y = "s06",
		]
		{avgHierarchyWidthPerLevel.csv};
		\end{axis}
  \end{tikzpicture}&%
%%%%%%%%%%%%%%%%%%%%%%%%%%%%%%%%%%%%%%%%%%%%%%%%%%%%%%
  \begin{tikzpicture}
    \begin{axis} 
		[
	  	ybar,
	  	title=s07,
		  xmin = -0.5,
		  xmax = 11.6,
		  bar width = 6.5,
		]
		\addplot
		[
			color = blue,
			fill = blue,
		]
		table
		[
	  	col sep = semicolon,
	  	x = "Level",
	  	y = "s07",
		]
		{avgHierarchyWidthPerLevel.csv};
		\end{axis}
  \end{tikzpicture}
%%%%%%%%%%%%%%%%%%%%%%%%%%%%%%%%%%%%%%%%%%%%%%%%%%%%%%
\end{tabular}
\caption{Average hierarchy width $(B)$ on every hierarchy level (number of nodes on every level) \textbf{without} execution of reassignment procedure. Vertical axes show hierarchy width and horizontal axes indicate hierarchy level.}\label{fig:level-by-level-hierarchy-width}
\end{figure*}
%%%%%%%%%%%%%%%%%%%%%%%%%%%%%%%%%%%%%%%%%%%%%%%%%%%%%%%%%%%%%%%%%%%%%%%%%%%%%%%%%%%%%%%%%%%%%%%%%%%%%%%%%%%%%%
\noindent
\begin{figure*}[p!]
\centering
\begin{tabular}{rrrr}
%%%%%%%%%%%%%%%%%%%%%%%%%%%%%%%%%%%%%%%%%%%%%%%%%%%%%
  \begin{tikzpicture}
    \begin{axis} 
		[
	  	ybar,
	  	title=s00,
		  xmin = -0.5,
		  xmax = 6.5,
		  bar width = 11.5,
		]
		\addplot
		[
			color = blue,
			fill = blue,
		]
		table
		[
	  	col sep = semicolon,
	  	x = "Level",
	  	y = "s00r",
		]
		{avgHierarchyWidthPerLevel.csv};
		\end{axis}
  \end{tikzpicture}
  &
%%%%%%%%%%%%%%%%%%%%%%%%%%%%%%%%%%%%%%%%%%%%%%%%%%%%%
  \begin{tikzpicture}
    \begin{axis} 
		[
	  	ybar,
	  	title=s01,
		  xmin = -0.6,
		  bar width = 4.1,
		]
		\addplot
		[
			color = blue,
			fill = blue,
		]
		table
		[
	  	col sep = semicolon,
	  	x = "Level",
	  	y = "s01r",
		]
		{avgHierarchyWidthPerLevel.csv};
		\end{axis}
  \end{tikzpicture}
  &
%%%%%%%%%%%%%%%%%%%%%%%%%%%%%%%%%%%%%%%%%%%%%%%%%%%%%
  \begin{tikzpicture}
    \begin{axis} 
		[
	  	ybar,
	  	title=s02,
		  xmin = -0.55,
		  bar width = 4.4,
		]
		\addplot
		[
			color = blue,
			fill = blue,
		]
		table
		[
	  	col sep = semicolon,
	  	x = "Level",
	  	y = "s02r",
		]
		{avgHierarchyWidthPerLevel.csv};
		\end{axis}
  \end{tikzpicture}
  &
%%%%%%%%%%%%%%%%%%%%%%%%%%%%%%%%%%%%%%%%%%%%%%%%%%%%%
  \begin{tikzpicture}
    \begin{axis} 
		[
	  	ybar,
	  	title=s03,
		  xmin = -0.5,
		  xmax = 8.5,
	   	bar width = 8.8,
		]
		\addplot
		[
			color = blue,
			fill = blue,
		]
		table
		[
	  	col sep = semicolon,
	  	x = "Level",
	  	y = "s03r",
		]
		{avgHierarchyWidthPerLevel.csv};
		\end{axis}
  \end{tikzpicture}
  \\
%%%%%%%%%%%%%%%%%%%%%%%%%%%%%%%%%%%%%%%%%%%%%%%%%%%%%
  \begin{tikzpicture}
    \begin{axis} 
		[
	  	ybar,
	  	title=s04,
		  xmin = -0.5,
		  bar width = 0.8,
		]
		\addplot
		[
			color = blue,
			fill = blue,
		]
		table
		[
	  	col sep = semicolon,
	  	x = "Level",
	  	y = "s04r",
		]
		{avgHierarchyWidthPerLevel.csv};
		\end{axis}
  \end{tikzpicture}
  &
%%%%%%%%%%%%%%%%%%%%%%%%%%%%%%%%%%%%%%%%%%%%%%%%%%%%%
  \begin{tikzpicture}
    \begin{axis} 
		[
	  	ybar,
	  	title=s05,
		  xmin = -0.45,
		  xmax = 9.3,
		  bar width = 8.1,
		]
		\addplot
		[
			color = blue,
			fill = blue,
		]
		table
		[
	  	col sep = semicolon,
	  	x = "Level",
	  	y = "s05r",
		]
		{avgHierarchyWidthPerLevel.csv};
		\end{axis}
  \end{tikzpicture}
  &
%%%%%%%%%%%%%%%%%%%%%%%%%%%%%%%%%%%%%%%%%%%%%%%%%%%%%
  \begin{tikzpicture}
    \begin{axis} 
		[
	  	ybar,
	  	title=s06,
		  xmin = -0.5,
		  xmax = 10.6,
		  bar width = 7.1,
		]
		\addplot
		[
			color = blue,
			fill = blue,
		]
		table
		[
	  	col sep = semicolon,
	  	x = "Level",
	  	y = "s06r",
		]
		{avgHierarchyWidthPerLevel.csv};
		\end{axis}
  \end{tikzpicture}
  &
  %%%%%%%%%%%%%%%%%%%%%%%%%%%%%%%%%%%%%%%%%%%%%%%%%%%%%%
  \begin{tikzpicture}
    \begin{axis} 
		[
	  	ybar,
	  	title=s07,
		  xmin = -0.5,
		  xmax = 11.6,
		  bar width = 6.5,
		]
		\addplot
		[
			color = blue,
			fill = blue,
		]
		table
		[
	  	col sep = semicolon,
	  	x = "Level",
	  	y = "s07r",
		]
		{avgHierarchyWidthPerLevel.csv};
		\end{axis}
  \end{tikzpicture}
%%%%%%%%%%%%%%%%%%%%%%%%%%%%%%%%%%%%%%%%%%%%%%%%%%%%%%
\end{tabular}
\caption{Average hierarchy width $(B)$ on every hierarchy level (number of nodes on every level) \textbf{with} execution of reassignment procedure. Vertical axes show hierarchy width and horizontal axes indicate hierarchy level.}\label{fig:level-by-level-hierarchy-width-R}
\end{figure*} %2
	\noindent
\begin{figure*}[p!]
\centering
\begin{tabular}{llll}
%%%%%%%%%%%%%%%%%%%%%%%%%%%%%%%%%%%%%%%%%%%%%%%%%%%%%
  \begin{tikzpicture}
    \begin{axis} 
		[
	  	ybar,
	  	title=s00,
		  xmin = -0.5,
		  xmax = 6.5,
		  bar width = 11.5,
		]
		\addplot
		[
			color = blue,
			fill = blue,
		]
		table
		[
	  	col sep = semicolon,
	  	x = "Level",
	  	y = "s00",
		]
		{avgNumOfInstancesPerLevel.csv};
		\end{axis}
  \end{tikzpicture}
  &
%%%%%%%%%%%%%%%%%%%%%%%%%%%%%%%%%%%%%%%%%%%%%%%%%%%%%
  \begin{tikzpicture}
    \begin{axis} 
		[
	  	ybar,
	  	title=s01,
		  xmin = -0.6,
		  bar width = 4.1,
		]
		\addplot
		[
			color = blue,
			fill = blue,
		]
		table
		[
	  	col sep = semicolon,
	  	x = "Level",
	  	y = "s01",
		]
		{avgNumOfInstancesPerLevel.csv};
		\end{axis}
  \end{tikzpicture}
  &
%%%%%%%%%%%%%%%%%%%%%%%%%%%%%%%%%%%%%%%%%%%%%%%%%%%%%
  \begin{tikzpicture}
    \begin{axis} 
		[
	  	ybar,
	  	title=s02,
		  xmin = -0.55,
		  bar width = 4.4,
		]
		\addplot
		[
			color = blue,
			fill = blue,
		]
		table
		[
	  	col sep = semicolon,
	  	x = "Level",
	  	y = "s02",
		]
		{avgNumOfInstancesPerLevel.csv};
		\end{axis}
  \end{tikzpicture}
  &
%%%%%%%%%%%%%%%%%%%%%%%%%%%%%%%%%%%%%%%%%%%%%%%%%%%%%
  \begin{tikzpicture}
    \begin{axis} 
		[
	  	ybar,
	  	title=s03,
		  xmin = -0.5,
		  xmax = 8.5,
	   	bar width = 8.8,
		]
		\addplot
		[
			color = blue,
			fill = blue,
		]
		table
		[
	  	col sep = semicolon,
	  	x = "Level",
	  	y = "s03",
		]
		{avgNumOfInstancesPerLevel.csv};
		\end{axis}
  \end{tikzpicture}
  \\
%%%%%%%%%%%%%%%%%%%%%%%%%%%%%%%%%%%%%%%%%%%%%%%%%%%%%
  \begin{tikzpicture}
    \begin{axis} 
		[
	  	ybar,
	  	title=s04,
		  xmin = -0.5,
		  bar width = 0.8,
		]
		\addplot
		[
			color = blue,
			fill = blue,
		]
		table
		[
	  	col sep = semicolon,
	  	x = "Level",
	  	y = "s04",
		]
		{avgNumOfInstancesPerLevel.csv};
		\end{axis}
  \end{tikzpicture}
  &
%%%%%%%%%%%%%%%%%%%%%%%%%%%%%%%%%%%%%%%%%%%%%%%%%%%%%
  \begin{tikzpicture}
    \begin{axis} 
		[
	  	ybar,
	  	title=s05,
		  xmin = -0.45,
		  xmax = 9.3,
		  bar width = 8.1,
		]
		\addplot
		[
			color = blue,
			fill = blue,
		]
		table
		[
	  	col sep = semicolon,
	  	x = "Level",
	  	y = "s05",
		]
		{avgNumOfInstancesPerLevel.csv};
		\end{axis}
  \end{tikzpicture}
  &
%%%%%%%%%%%%%%%%%%%%%%%%%%%%%%%%%%%%%%%%%%%%%%%%%%%%%
  \begin{tikzpicture}
    \begin{axis} 
		[
	  	ybar,
	  	title=s06,
		  xmin = -0.5,
		  xmax = 10.6,
		  bar width = 7.5,
		]
		\addplot
		[
			color = blue,
			fill = blue,
		]
		table
		[
	  	col sep = semicolon,
	  	x = "Level",
	  	y = "s06",
		]
		{avgNumOfInstancesPerLevel.csv};
		\end{axis}
  \end{tikzpicture}
  &
%%%%%%%%%%%%%%%%%%%%%%%%%%%%%%%%%%%%%%%%%%%%%%%%%%%%%%
  \begin{tikzpicture}
    \begin{axis} 
		[
	  	ybar,
	  	title=s07,
		  xmin = -0.5,
		  xmax = 11.6,
		  bar width = 6.8,
		]
		\addplot
		[
			color = blue,
			fill = blue,
		]
		table
		[
	  	col sep = semicolon,
	  	x = "Level",
	  	y = "s07",
		]
		{avgNumOfInstancesPerLevel.csv};
		\end{axis}
  \end{tikzpicture}
%%%%%%%%%%%%%%%%%%%%%%%%%%%%%%%%%%%%%%%%%%%%%%%%%%%%%%
\end{tabular}
\caption{Average distribution of data instances among hierarchy levels \textbf{without} execution of reassignment procedure. Vertical axes show the number of instances and horizontal axes indicate hierarchy level.}\label{fig:level-by-level-number-of-instances}
\end{figure*}
%%%%%%%%%%%%%%%%%%%%%%%%%%%%%%%%%%%%%%%%%%%%%%%%%%%%%%%%%%%%%%%%%%%%%%%%%%%%%%%%%%%%%%%%%%%%%%%%%%%%%%%%%%%%%%
\noindent
\begin{figure*}[p!]
\centering
\begin{tabular}{llll}
%%%%%%%%%%%%%%%%%%%%%%%%%%%%%%%%%%%%%%%%%%%%%%%%%%%%%
  \begin{tikzpicture}
    \begin{axis} 
		[
	  	ybar,
	  	title=s00,
		  xmin = -0.5,
		  xmax = 6.5,
		  bar width = 11.6,
		]
		\addplot
		[
			color = blue,
			fill = blue,
		]
		table
		[
	  	col sep = semicolon,
	  	x = "Level",
	  	y = "s00r",
		]
		{avgNumOfInstancesPerLevel.csv};
		\end{axis}
  \end{tikzpicture}
  &
%%%%%%%%%%%%%%%%%%%%%%%%%%%%%%%%%%%%%%%%%%%%%%%%%%%%%
  \begin{tikzpicture}
    \begin{axis} 
		[
	  	ybar,
	  	title=s01,
		  xmin = -0.6,
		  bar width = 4.1,
		]
		\addplot
		[
			color = blue,
			fill = blue,
		]
		table
		[
	  	col sep = semicolon,
	  	x = "Level",
	  	y = "s01r",
		]
		{avgNumOfInstancesPerLevel.csv};
		\end{axis}
  \end{tikzpicture}
  &
%%%%%%%%%%%%%%%%%%%%%%%%%%%%%%%%%%%%%%%%%%%%%%%%%%%%%
  \begin{tikzpicture}
    \begin{axis} 
		[
	  	ybar,
	  	title=s02,
		  xmin = -0.55,
		  bar width = 4.4,
		]
		\addplot
		[
			color = blue,
			fill = blue,
		]
		table
		[
	  	col sep = semicolon,
	  	x = "Level",
	  	y = "s02r",
		]
		{avgNumOfInstancesPerLevel.csv};
		\end{axis}
  \end{tikzpicture}
  &
%%%%%%%%%%%%%%%%%%%%%%%%%%%%%%%%%%%%%%%%%%%%%%%%%%%%%
  \begin{tikzpicture}
    \begin{axis} 
		[
	  	ybar,
	  	title=s03,
		  xmin = -0.5,
		  xmax = 8.5,
	   	bar width = 9.0,
		]
		\addplot
		[
			color = blue,
			fill = blue,
		]
		table
		[
	  	col sep = semicolon,
	  	x = "Level",
	  	y = "s03r",
		]
		{avgNumOfInstancesPerLevel.csv};
		\end{axis}
  \end{tikzpicture}
  \\
%%%%%%%%%%%%%%%%%%%%%%%%%%%%%%%%%%%%%%%%%%%%%%%%%%%%%
  \begin{tikzpicture}
    \begin{axis} 
		[
	  	ybar,
	  	title=s04,
		  xmin = -0.5,
		  bar width = 0.8,
		]
		\addplot
		[
			color = blue,
			fill = blue,
		]
		table
		[
	  	col sep = semicolon,
	  	x = "Level",
	  	y = "s04r",
		]
		{avgNumOfInstancesPerLevel.csv};
		\end{axis}
  \end{tikzpicture}
  &
%%%%%%%%%%%%%%%%%%%%%%%%%%%%%%%%%%%%%%%%%%%%%%%%%%%%%
  \begin{tikzpicture}
    \begin{axis} 
		[
	  	ybar,
	  	title=s05,
		  xmin = -0.45,
		  xmax = 9.3,
		  bar width = 8.1,
		]
		\addplot
		[
			color = blue,
			fill = blue,
		]
		table
		[
	  	col sep = semicolon,
	  	x = "Level",
	  	y = "s05r",
		]
		{avgNumOfInstancesPerLevel.csv};
		\end{axis}
  \end{tikzpicture}
  &
%%%%%%%%%%%%%%%%%%%%%%%%%%%%%%%%%%%%%%%%%%%%%%%%%%%%%
  \begin{tikzpicture}
    \begin{axis} 
		[
	  	ybar,
	  	title=s06,
		  xmin = -0.5,
		  xmax = 10.6,
		  bar width = 7.1,
		]
		\addplot
		[
			color = blue,
			fill = blue,
		]
		table
		[
	  	col sep = semicolon,
	  	x = "Level",
	  	y = "s06r",
		]
		{avgNumOfInstancesPerLevel.csv};
		\end{axis}
  \end{tikzpicture}
  &
%%%%%%%%%%%%%%%%%%%%%%%%%%%%%%%%%%%%%%%%%%%%%%%%%%%%%%
  \begin{tikzpicture}
    \begin{axis} 
		[
	  	ybar,
	  	title=s07,
		  xmin = -0.5,
		  xmax = 11.6,
		  bar width = 6.5,
		]
		\addplot
		[
			color = blue,
			fill = blue,
		]
		table
		[
	  	col sep = semicolon,
	  	x = "Level",
	  	y = "s07r",
		]
		{avgNumOfInstancesPerLevel.csv};
		\end{axis}
  \end{tikzpicture}
%%%%%%%%%%%%%%%%%%%%%%%%%%%%%%%%%%%%%%%%%%%%%%%%%%%%%%
\end{tabular}
\caption{Average distribution of data instances among hierarchy levels \textbf{with} execution of reassignment procedure. Vertical axes show the number of instances and horizontal axes indicate hierarchy level.}\label{fig:level-by-level-number-of-instances-R}
\end{figure*} %1
	\noindent
\begin{figure*}[p!]
\centering
\begin{tabular}{rrrr}
%%%%%%%%%%%%%%%%%%%%%%%%%%%%%%%%%%%%%%%%%%%%%%%%%%%%%
  \begin{tikzpicture}
    \begin{axis} 
		[
	  	ybar,
	  	title=s00,
		  xmin = -0.5,
		  xmax = 6.5,
		  bar width = 11.6,
		]
		\addplot
		[
			color = blue,
			fill = blue,
		]
		table
		[
	  	col sep = semicolon,
	  	x = "Level",
	  	y = "s00",
		]
		{avgOfAvgNumOfChildrenPerNodePerLevel.csv};
		\end{axis}
  \end{tikzpicture}
  &
%%%%%%%%%%%%%%%%%%%%%%%%%%%%%%%%%%%%%%%%%%%%%%%%%%%%%
  \begin{tikzpicture}
    \begin{axis} 
		[
	  	ybar,
	  	title=s01,
		  xmin = -0.6,
		  bar width = 4.2,
		]
		\addplot
		[
			color = blue,
			fill = blue,
		]
		table
		[
	  	col sep = semicolon,
	  	x = "Level",
	  	y = "s01",
		]
		{avgOfAvgNumOfChildrenPerNodePerLevel.csv};
		\end{axis}
  \end{tikzpicture}
  &
%%%%%%%%%%%%%%%%%%%%%%%%%%%%%%%%%%%%%%%%%%%%%%%%%%%%%
  \begin{tikzpicture}
    \begin{axis} 
		[
	  	ybar,
	  	title=s02,
		  xmin = -0.55,
		  bar width = 4.4,
		]
		\addplot
		[
			color = blue,
			fill = blue,
		]
		table
		[
	  	col sep = semicolon,
	  	x = "Level",
	  	y = "s02",
		]
		{avgOfAvgNumOfChildrenPerNodePerLevel.csv};
		\end{axis}
  \end{tikzpicture}
  &
%%%%%%%%%%%%%%%%%%%%%%%%%%%%%%%%%%%%%%%%%%%%%%%%%%%%%
  \begin{tikzpicture}
    \begin{axis} 
		[
	  	ybar,
	  	title=s03,
		  xmin = -0.5,
		  xmax = 8.5,
	   	bar width = 8.9,
		]
		\addplot
		[
			color = blue,
			fill = blue,
		]
		table
		[
	  	col sep = semicolon,
	  	x = "Level",
	  	y = "s03",
		]
		{avgOfAvgNumOfChildrenPerNodePerLevel.csv};
		\end{axis}
  \end{tikzpicture}
  \\
%%%%%%%%%%%%%%%%%%%%%%%%%%%%%%%%%%%%%%%%%%%%%%%%%%%%%
  \begin{tikzpicture}
    \begin{axis} 
		[
	  	ybar,
	  	title=s04,
		  xmin = -0.5,
		  bar width = 0.8,
		]
		\addplot
		[
			color = blue,
			fill = blue,
		]
		table
		[
	  	col sep = semicolon,
	  	x = "Level",
	  	y = "s04",
		]
		{avgOfAvgNumOfChildrenPerNodePerLevel.csv};
		\end{axis}
  \end{tikzpicture}
  &
%%%%%%%%%%%%%%%%%%%%%%%%%%%%%%%%%%%%%%%%%%%%%%%%%%%%%
  \begin{tikzpicture}
    \begin{axis} 
		[
	  	ybar,
	  	title=s05,
		  xmin = -0.45,
		  xmax = 9.3,
		  bar width = 8.5,
		]
		\addplot
		[
			color = blue,
			fill = blue,
		]
		table
		[
	  	col sep = semicolon,
	  	x = "Level",
	  	y = "s05",
		]
		{avgOfAvgNumOfChildrenPerNodePerLevel.csv};
		\end{axis}
  \end{tikzpicture}
  &
%%%%%%%%%%%%%%%%%%%%%%%%%%%%%%%%%%%%%%%%%%%%%%%%%%%%%
  \begin{tikzpicture}
    \begin{axis} 
		[
	  	ybar,
	  	title=s06,
		  xmin = -0.5,
		  xmax = 10.6,
		  bar width = 7.1,
		]
		\addplot
		[
			color = blue,
			fill = blue,
		]
		table
		[
	  	col sep = semicolon,
	  	x = "Level",
	  	y = "s06",
		]
		{avgOfAvgNumOfChildrenPerNodePerLevel.csv};
		\end{axis}
  \end{tikzpicture}
  &
%%%%%%%%%%%%%%%%%%%%%%%%%%%%%%%%%%%%%%%%%%%%%%%%%%%%%%
  \begin{tikzpicture}
    \begin{axis} 
		[
	  	ybar,
	  	title=s07,
		  xmin = -0.5,
		  xmax = 11.6,
		  bar width = 6.5,
		]
		\addplot
		[
			color = blue,
			fill = blue,
		]
		table
		[
	  	col sep = semicolon,
	  	x = "Level",
	  	y = "s07",
		]
		{avgOfAvgNumOfChildrenPerNodePerLevel.csv};
		\end{axis}
  \end{tikzpicture}
%%%%%%%%%%%%%%%%%%%%%%%%%%%%%%%%%%%%%%%%%%%%%%%%%%%%%%
\end{tabular}
\caption{Distribution of the average number of children per node among hierarchy levels \textbf{without} execution of reassignment procedure. Vertical axes show the number of children and horizontal axes indicate hierarchy level.}\label{fig:level-by-level-number-of-children-per-node}
\end{figure*}
%%%%%%%%%%%%%%%%%%%%%%%%%%%%%%%%%%%%%%%%%%%%%%%%%%%%%%%%%%%%%%%%%%%%%%%%%%%%%%%%%%%%%%%%%%%%%%%%%%%%%%%%%%%%%%
\noindent
\begin{figure*}[p!]
\centering
\begin{tabular}{rrrr}
%%%%%%%%%%%%%%%%%%%%%%%%%%%%%%%%%%%%%%%%%%%%%%%%%%%%%
  \begin{tikzpicture}
    \begin{axis} 
		[
	  	ybar,
	  	title=s00,
		  xmin = -0.5,
		  xmax = 6.5,
		  bar width = 11.5,
		]
		\addplot
		[
			color = blue,
			fill = blue,
		]
		table
		[
	  	col sep = semicolon,
	  	x = "Level",
	  	y = "s00r",
		]
		{avgOfAvgNumOfChildrenPerNodePerLevel.csv};
		\end{axis}
  \end{tikzpicture}
  &
%%%%%%%%%%%%%%%%%%%%%%%%%%%%%%%%%%%%%%%%%%%%%%%%%%%%%
  \begin{tikzpicture}
    \begin{axis} 
		[
	  	ybar,
	  	title=s01,
		  xmin = -0.6,
		  bar width = 4.9,
		]
		\addplot
		[
			color = blue,
			fill = blue,
		]
		table
		[
	  	col sep = semicolon,
	  	x = "Level",
	  	y = "s01r",
		]
		{avgOfAvgNumOfChildrenPerNodePerLevel.csv};
		\end{axis}
  \end{tikzpicture}
  &
%%%%%%%%%%%%%%%%%%%%%%%%%%%%%%%%%%%%%%%%%%%%%%%%%%%%%
  \begin{tikzpicture}
    \begin{axis} 
		[
	  	ybar,
	  	title=s02,
		  xmin = -0.55,
		  bar width = 5.0,
		]
		\addplot
		[
			color = blue,
			fill = blue,
		]
		table
		[
	  	col sep = semicolon,
	  	x = "Level",
	  	y = "s02r",
		]
		{avgOfAvgNumOfChildrenPerNodePerLevel.csv};
		\end{axis}
  \end{tikzpicture}
  &
%%%%%%%%%%%%%%%%%%%%%%%%%%%%%%%%%%%%%%%%%%%%%%%%%%%%%
  \begin{tikzpicture}
    \begin{axis} 
		[
	  	ybar,
	  	title=s03,
		  xmin = -0.5,
		  xmax = 8.5,
	   	bar width = 8.8,
		]
		\addplot
		[
			color = blue,
			fill = blue,
		]
		table
		[
	  	col sep = semicolon,
	  	x = "Level",
	  	y = "s03r",
		]
		{avgOfAvgNumOfChildrenPerNodePerLevel.csv};
		\end{axis}
  \end{tikzpicture}
  \\
%%%%%%%%%%%%%%%%%%%%%%%%%%%%%%%%%%%%%%%%%%%%%%%%%%%%%
  \begin{tikzpicture}
    \begin{axis} 
		[
	  	ybar,
	  	title=s04,
		  xmin = -0.5,
		  bar width = 0.8,
		]
		\addplot
		[
			color = blue,
			fill = blue,
		]
		table
		[
	  	col sep = semicolon,
	  	x = "Level",
	  	y = "s04r",
		]
		{avgOfAvgNumOfChildrenPerNodePerLevel.csv};
		\end{axis}
  \end{tikzpicture}
  &
%%%%%%%%%%%%%%%%%%%%%%%%%%%%%%%%%%%%%%%%%%%%%%%%%%%%%
  \begin{tikzpicture}
    \begin{axis} 
		[
	  	ybar,
	  	title=s05,
		  xmin = -0.45,
		  xmax = 9.3,
		  bar width = 8.1,
		]
		\addplot
		[
			color = blue,
			fill = blue,
		]
		table
		[
	  	col sep = semicolon,
	  	x = "Level",
	  	y = "s05r",
		]
		{avgOfAvgNumOfChildrenPerNodePerLevel.csv};
		\end{axis}
  \end{tikzpicture}
  &
%%%%%%%%%%%%%%%%%%%%%%%%%%%%%%%%%%%%%%%%%%%%%%%%%%%%%
  \begin{tikzpicture}
    \begin{axis} 
		[
	  	ybar,
	  	title=s06,
		  xmin = -0.5,
		  xmax = 10.6,
		  bar width = 7.1,
		]
		\addplot
		[
			color = blue,
			fill = blue,
		]
		table
		[
	  	col sep = semicolon,
	  	x = "Level",
	  	y = "s06r",
		]
		{avgOfAvgNumOfChildrenPerNodePerLevel.csv};
		\end{axis}
  \end{tikzpicture}
  &
%%%%%%%%%%%%%%%%%%%%%%%%%%%%%%%%%%%%%%%%%%%%%%%%%%%%%%
  \begin{tikzpicture}
    \begin{axis} 
		[
	  	ybar,
	  	title=s07,
		  xmin = -0.5,
		  xmax = 11.6,
		  bar width = 6.8,
		]
		\addplot
		[
			color = blue,
			fill = blue,
		]
		table
		[
	  	col sep = semicolon,
	  	x = "Level",
	  	y = "s07r",
		]
		{avgOfAvgNumOfChildrenPerNodePerLevel.csv};
		\end{axis}
  \end{tikzpicture}
%%%%%%%%%%%%%%%%%%%%%%%%%%%%%%%%%%%%%%%%%%%%%%%%%%%%%%
\end{tabular}
\caption{Distribution of the average number of children per node among hierarchy levels \textbf{with} execution of reassignment procedure. Vertical axes show the number of children and horizontal axes indicate hierarchy level.}\label{fig:level-by-level-number-of-children-per-node-R}
\end{figure*} %3
	\noindent
\begin{figure*}[p!]
\centering
\begin{tabular}{rrrr}
%%%%%%%%%%%%%%%%%%%%%%%%%%%%%%%%%%%%%%%%%%%%%%%%%%%%%
  \begin{tikzpicture}
    \begin{axis} 
		[
	  	ybar,
	  	title=s00,
		  xmin = -0.5,
		  xmax = 6.5,
		  bar width = 11.5,
		]
		\addplot
		[
			color = blue,
			fill = blue,
		]
		table
		[
	  	col sep = semicolon,
	  	x = "Level",
	  	y = "s00",
		]
		{avgNumOfLeavesPerLevel.csv};
		\end{axis}
  \end{tikzpicture}
  &
%%%%%%%%%%%%%%%%%%%%%%%%%%%%%%%%%%%%%%%%%%%%%%%%%%%%%
  \begin{tikzpicture}
    \begin{axis} 
		[
	  	ybar,
	  	title=s01,
		  xmin = -0.6,
		  bar width = 4.1,
		]
		\addplot
		[
			color = blue,
			fill = blue,
		]
		table
		[
	  	col sep = semicolon,
	  	x = "Level",
	  	y = "s01",
		]
		{avgNumOfLeavesPerLevel.csv};
		\end{axis}
  \end{tikzpicture}
  &
%%%%%%%%%%%%%%%%%%%%%%%%%%%%%%%%%%%%%%%%%%%%%%%%%%%%%
  \begin{tikzpicture}
    \begin{axis} 
		[
	  	ybar,
	  	title=s02,
		  xmin = -0.55,
		  bar width = 4.4,
		]
		\addplot
		[
			color = blue,
			fill = blue,
		]
		table
		[
	  	col sep = semicolon,
	  	x = "Level",
	  	y = "s02",
		]
		{avgNumOfLeavesPerLevel.csv};
		\end{axis}
  \end{tikzpicture}
  &
%%%%%%%%%%%%%%%%%%%%%%%%%%%%%%%%%%%%%%%%%%%%%%%%%%%%%
  \begin{tikzpicture}
    \begin{axis} 
		[
	  	ybar,
	  	title=s03,
		  xmin = -0.5,
		  xmax = 8.5,
	   	bar width = 8.8,
		]
		\addplot
		[
			color = blue,
			fill = blue,
		]
		table
		[
	  	col sep = semicolon,
	  	x = "Level",
	  	y = "s03",
		]
		{avgNumOfLeavesPerLevel.csv};
		\end{axis}
  \end{tikzpicture}
  \\
%%%%%%%%%%%%%%%%%%%%%%%%%%%%%%%%%%%%%%%%%%%%%%%%%%%%%
  \begin{tikzpicture}
    \begin{axis} 
		[
	  	ybar,
	  	title=s04,
		  xmin = -0.5,
		  bar width = 0.8,
		]
		\addplot
		[
			color = blue,
			fill = blue,
		]
		table
		[
	  	col sep = semicolon,
	  	x = "Level",
	  	y = "s04",
		]
		{avgNumOfLeavesPerLevel.csv};
		\end{axis}
  \end{tikzpicture}
  &
%%%%%%%%%%%%%%%%%%%%%%%%%%%%%%%%%%%%%%%%%%%%%%%%%%%%%
  \begin{tikzpicture}
    \begin{axis} 
		[
	  	ybar,
	  	title=s05,
		  xmin = -0.45,
		  xmax = 9.3,
		  bar width = 8.2,
		]
		\addplot
		[
			color = blue,
			fill = blue,
		]
		table
		[
	  	col sep = semicolon,
	  	x = "Level",
	  	y = "s05",
		]
		{avgNumOfLeavesPerLevel.csv};
		\end{axis}
  \end{tikzpicture}
  &
%%%%%%%%%%%%%%%%%%%%%%%%%%%%%%%%%%%%%%%%%%%%%%%%%%%%%
  \begin{tikzpicture}
    \begin{axis} 
		[
	  	ybar,
	  	title=s06,
		  xmin = -0.5,
		  xmax = 10.6,
		  bar width = 7.1,
		]
		\addplot
		[
			color = blue,
			fill = blue,
		]
		table
		[
	  	col sep = semicolon,
	  	x = "Level",
	  	y = "s06",
		]
		{avgNumOfLeavesPerLevel.csv};
		\end{axis}
  \end{tikzpicture}
  &
%%%%%%%%%%%%%%%%%%%%%%%%%%%%%%%%%%%%%%%%%%%%%%%%%%%%%%
  \begin{tikzpicture}
    \begin{axis} 
		[
	  	ybar,
	  	title=s07,
		  xmin = -0.5,
		  xmax = 11.6,
		  bar width = 6.5,
		]
		\addplot
		[
			color = blue,
			fill = blue,
		]
		table
		[
	  	col sep = semicolon,
	  	x = "Level",
	  	y = "s07",
		]
		{avgNumOfLeavesPerLevel.csv};
		\end{axis}
  \end{tikzpicture}
%%%%%%%%%%%%%%%%%%%%%%%%%%%%%%%%%%%%%%%%%%%%%%%%%%%%%%
\end{tabular}
\caption{Average number of leaf nodes $(L)$ on every hierarchy level \textbf{without} execution of reassignment procedure. Vertical axes show the number of children and horizontal axes indicate hierarchy level.}\label{fig:level-by-level-number-of-leaves}
\end{figure*}
%%%%%%%%%%%%%%%%%%%%%%%%%%%%%%%%%%%%%%%%%%%%%%%%%%%%%%%%%%%%%%%%%%%%%%%%%%%%%%%%%%%%%%%%%%%%%%%%%%%%%%%%%%%%%%
\noindent
\begin{figure*}[p!]
\centering
\begin{tabular}{rrrr}
%%%%%%%%%%%%%%%%%%%%%%%%%%%%%%%%%%%%%%%%%%%%%%%%%%%%%
  \begin{tikzpicture}
    \begin{axis} 
		[
	  	ybar,
	  	title=s00,
		  xmin = -0.5,
		  xmax = 6.5,
		  bar width = 11.5,
		]
		\addplot
		[
			color = blue,
			fill = blue,
		]
		table
		[
	  	col sep = semicolon,
	  	x = "Level",
	  	y = "s00r",
		]
		{avgNumOfLeavesPerLevel.csv};
		\end{axis}
  \end{tikzpicture}
  &
%%%%%%%%%%%%%%%%%%%%%%%%%%%%%%%%%%%%%%%%%%%%%%%%%%%%%
  \begin{tikzpicture}
    \begin{axis} 
		[
	  	ybar,
	  	title=s01,
		  xmin = -0.6,
		  bar width = 4.1,
		]
		\addplot
		[
			color = blue,
			fill = blue,
		]
		table
		[
	  	col sep = semicolon,
	  	x = "Level",
	  	y = "s01r",
		]
		{avgNumOfLeavesPerLevel.csv};
		\end{axis}
  \end{tikzpicture}
  &
%%%%%%%%%%%%%%%%%%%%%%%%%%%%%%%%%%%%%%%%%%%%%%%%%%%%%
  \begin{tikzpicture}
    \begin{axis} 
		[
	  	ybar,
	  	title=s02,
		  xmin = -0.55,
		  bar width = 4.4,
		]
		\addplot
		[
			color = blue,
			fill = blue,
		]
		table
		[
	  	col sep = semicolon,
	  	x = "Level",
	  	y = "s02r",
		]
		{avgNumOfLeavesPerLevel.csv};
		\end{axis}
  \end{tikzpicture}
  &
%%%%%%%%%%%%%%%%%%%%%%%%%%%%%%%%%%%%%%%%%%%%%%%%%%%%%
  \begin{tikzpicture}
    \begin{axis} 
		[
	  	ybar,
	  	title=s03,
		  xmin = -0.5,
		  xmax = 8.5,
	   	bar width = 8.8,
		]
		\addplot
		[
			color = blue,
			fill = blue,
		]
		table
		[
	  	col sep = semicolon,
	  	x = "Level",
	  	y = "s03r",
		]
		{avgNumOfLeavesPerLevel.csv};
		\end{axis}
  \end{tikzpicture}
  \\
%%%%%%%%%%%%%%%%%%%%%%%%%%%%%%%%%%%%%%%%%%%%%%%%%%%%%
  \begin{tikzpicture}
    \begin{axis} 
		[
	  	ybar,
	  	title=s04,
		  xmin = -0.5,
		  bar width = 0.8,
		]
		\addplot
		[
			color = blue,
			fill = blue,
		]
		table
		[
	  	col sep = semicolon,
	  	x = "Level",
	  	y = "s04r",
		]
		{avgNumOfLeavesPerLevel.csv};
		\end{axis}
  \end{tikzpicture}
  &
%%%%%%%%%%%%%%%%%%%%%%%%%%%%%%%%%%%%%%%%%%%%%%%%%%%%%
  \begin{tikzpicture}
    \begin{axis} 
		[
	  	ybar,
	  	title=s05,
		  xmin = -0.45,
		  xmax = 9.3,
		  bar width = 8.2,
		]
		\addplot
		[
			color = blue,
			fill = blue,
		]
		table
		[
	  	col sep = semicolon,
	  	x = "Level",
	  	y = "s05r",
		]
		{avgNumOfLeavesPerLevel.csv};
		\end{axis}
  \end{tikzpicture}
  &
%%%%%%%%%%%%%%%%%%%%%%%%%%%%%%%%%%%%%%%%%%%%%%%%%%%%%
  \begin{tikzpicture}
    \begin{axis} 
		[
	  	ybar,
	  	title=s06,
		  xmin = -0.5,
		  xmax = 10.6,
		  bar width = 7.1,
		]
		\addplot
		[
			color = blue,
			fill = blue,
		]
		table
		[
	  	col sep = semicolon,
	  	x = "Level",
	  	y = "s06r",
		]
		{avgNumOfLeavesPerLevel.csv};
		\end{axis}
  \end{tikzpicture}
  &
%%%%%%%%%%%%%%%%%%%%%%%%%%%%%%%%%%%%%%%%%%%%%%%%%%%%%%
  \begin{tikzpicture}
    \begin{axis} 
		[
	  	ybar,
	  	title=s07,
		  xmin = -0.5,
		  xmax = 11.6,
		  bar width = 6.5,
		]
		\addplot
		[
			color = blue,
			fill = blue,
		]
		table
		[
	  	col sep = semicolon,
	  	x = "Level",
	  	y = "s07r",
		]
		{avgNumOfLeavesPerLevel.csv};
		\end{axis}
  \end{tikzpicture}
%%%%%%%%%%%%%%%%%%%%%%%%%%%%%%%%%%%%%%%%%%%%%%%%%%%%%%
\end{tabular}
\caption{Average number of leaf nodes $(L)$ on every hierarchy level \textbf{with} execution of reassignment procedure. Vertical axes show the number of children and horizontal axes indicate hierarchy level.}\label{fig:level-by-level-number-of-leaves-R}
\end{figure*} %4
	\noindent
\begin{figure*}[p!]
\centering
\begin{tabular}{rrrr}
%%%%%%%%%%%%%%%%%%%%%%%%%%%%%%%%%%%%%%%%%%%%%%%%%%%%%
  \begin{tikzpicture}
    \begin{axis} 
		[
	  	ybar,
	  	title=s00,
		  xmin = 0.5,
		  xmax = 9.5,
		  bar width = 8.9,
		]
		\addplot
		[
			color = blue,
			fill = blue,
		]
		table
		[
	  	col sep = semicolon,
	  	x = "Factor",
	  	y = "s00",
		]
		{branchingFactorHistogram.csv};
		\end{axis}
  \end{tikzpicture}
  &
%%%%%%%%%%%%%%%%%%%%%%%%%%%%%%%%%%%%%%%%%%%%%%%%%%%%%
  \begin{tikzpicture}
    \begin{axis} 
		[
	  	ybar,
	  	title=s01,
		  xmin = 0.6,
		  xmax = 8.5,
		  bar width = 10.1,
		]
		\addplot
		[
			color = blue,
			fill = blue,
		]
		table
		[
	  	col sep = semicolon,
	  	x = "Factor",
	  	y = "s01",
		]
		{branchingFactorHistogram.csv};
		\end{axis}
  \end{tikzpicture}
  &
%%%%%%%%%%%%%%%%%%%%%%%%%%%%%%%%%%%%%%%%%%%%%%%%%%%%%
  \begin{tikzpicture}
    \begin{axis} 
		[
	  	ybar,
	  	title=s02,
		  xmin = 0.45,
		  xmax = 19.0,
		  bar width = 4.1,
		]
		\addplot
		[
			color = blue,
			fill = blue,
		]
		table
		[
	  	col sep = semicolon,
	  	x = "Factor",
	  	y = "s02",
		]
		{branchingFactorHistogram.csv};
		\end{axis}
  \end{tikzpicture}
  &
%%%%%%%%%%%%%%%%%%%%%%%%%%%%%%%%%%%%%%%%%%%%%%%%%%%%%
  \begin{tikzpicture}
    \begin{axis} 
		[
	  	ybar,
	  	title=s03,
		  xmin = 0.5,
		  xmax = 12,
	   	bar width = 6.9,
		]
		\addplot
		[
			color = blue,
			fill = blue,
		]
		table
		[
	  	col sep = semicolon,
	  	x = "Factor",
	  	y = "s03",
		]
		{branchingFactorHistogram.csv};
		\end{axis}
  \end{tikzpicture}
  \\
%%%%%%%%%%%%%%%%%%%%%%%%%%%%%%%%%%%%%%%%%%%%%%%%%%%%%
  \begin{tikzpicture}
    \begin{axis} 
		[
	  	ybar,
	  	title=s04,
		  xmin = 0.5,
		  xmax = 12.5,
		  bar width = 6.5,
		]
		\addplot
		[
			color = blue,
			fill = blue,
		]
		table
		[
	  	col sep = semicolon,
	  	x = "Factor",
	  	y = "s04",
		]
		{branchingFactorHistogram.csv};
		\end{axis}
  \end{tikzpicture}
  &
%%%%%%%%%%%%%%%%%%%%%%%%%%%%%%%%%%%%%%%%%%%%%%%%%%%%%
  \begin{tikzpicture}
    \begin{axis} 
		[
	  	ybar,
	  	title=s05,
		  xmin = 0.5,
		  xmax = 20.0,
		  bar width = 3.8,
		]
		\addplot
		[
			color = blue,
			fill = blue,
		]
		table
		[
	  	col sep = semicolon,
	  	x = "Factor",
	  	y = "s05",
		]
		{branchingFactorHistogram.csv};
		\end{axis}
  \end{tikzpicture}
  &
%%%%%%%%%%%%%%%%%%%%%%%%%%%%%%%%%%%%%%%%%%%%%%%%%%%%%
  \begin{tikzpicture}
    \begin{axis} 
		[
	  	ybar,
	  	title=s06,
		  xmin = 0.5,
		  xmax = 10,
		  bar width = 8.28,
		]
		\addplot
		[
			color = blue,
			fill = blue,
		]
		table
		[
	  	col sep = semicolon,
	  	x = "Factor",
	  	y = "s06",
		]
		{branchingFactorHistogram.csv};
		\end{axis}
  \end{tikzpicture}
  &
%%%%%%%%%%%%%%%%%%%%%%%%%%%%%%%%%%%%%%%%%%%%%%%%%%%%%%
  \begin{tikzpicture}
    \begin{axis} 
		[
	  	ybar,
	  	title=s07,
		  xmin = 0.4,
		  xmax = 24.6,
		  bar width = 3.2,
		]
		\addplot
		[
			color = blue,
			fill = blue,
		]
		table
		[
	  	col sep = semicolon,
	  	x = "Factor",
	  	y = "s07",
		]
		{branchingFactorHistogram.csv};
		\end{axis}
  \end{tikzpicture}%
%%%%%%%%%%%%%%%%%%%%%%%%%%%%%%%%%%%%%%%%%%%%%%%%%%%%%%
\end{tabular}
\caption{Average number of child nodes for every node in generated hierarchies \textbf{without} execution of reassignment procedure. Horizontal axes show the number of children and vertical axes show the number of occurrences (count) in the hierarchies.}\label{fig:branching-factor-histogram}
\end{figure*}
%%%%%%%%%%%%%%%%%%%%%%%%%%%%%%%%%%%%%%%%%%%%%%%%%%%%%%%%%%%%%%%%%%%%%%%%%%%%%%%%%%%%%%%%%%%%%%%%%%%%%%%%%%%%%%
\noindent    
\begin{figure*}[p!]
\centering
\begin{tabular}{rrrr}
%%%%%%%%%%%%%%%%%%%%%%%%%%%%%%%%%%%%%%%%%%%%%%%%%%%%%
  \begin{tikzpicture}
    \begin{axis} 
		[
	  	ybar,
	  	title=s00,
		  xmin = 0.5,
		  xmax = 9.5,
		  bar width = 8.9,
		]
		\addplot
		[
			color = blue,
			fill = blue,
		]
		table
		[
	  	col sep = semicolon,
	  	x = "Factor",
	  	y = "s00r",
		]
		{branchingFactorHistogram.csv};
		\end{axis}
  \end{tikzpicture}
  &
%%%%%%%%%%%%%%%%%%%%%%%%%%%%%%%%%%%%%%%%%%%%%%%%%%%%%
  \begin{tikzpicture}
    \begin{axis} 
		[
	  	ybar,
	  	title=s01,
		  xmin = 0.6,
		  xmax = 8.5,
		  bar width = 10.1,
		]
		\addplot
		[
			color = blue,
			fill = blue,
		]
		table
		[
	  	col sep = semicolon,
	  	x = "Factor",
	  	y = "s01r",
		]
		{branchingFactorHistogram.csv};
		\end{axis}
  \end{tikzpicture}
  &
%%%%%%%%%%%%%%%%%%%%%%%%%%%%%%%%%%%%%%%%%%%%%%%%%%%%%
  \begin{tikzpicture}
    \begin{axis} 
		[
	  	ybar,
	  	title=s02,
		  xmin = 0.45,
		  xmax = 19.0,
		  bar width = 4.1,
		]
		\addplot
		[
			color = blue,
			fill = blue,
		]
		table
		[
	  	col sep = semicolon,
	  	x = "Factor",
	  	y = "s02r",
		]
		{branchingFactorHistogram.csv};
		\end{axis}
  \end{tikzpicture}
  &
%%%%%%%%%%%%%%%%%%%%%%%%%%%%%%%%%%%%%%%%%%%%%%%%%%%%%
  \begin{tikzpicture}
    \begin{axis} 
		[
	  	ybar,
	  	title=s03,
		  xmin = 0.5,
		  xmax = 12,
	   	bar width = 6.9,
		]
		\addplot
		[
			color = blue,
			fill = blue,
		]
		table
		[
	  	col sep = semicolon,
	  	x = "Factor",
	  	y = "s03r",
		]
		{branchingFactorHistogram.csv};
		\end{axis}
  \end{tikzpicture}
  \\
%%%%%%%%%%%%%%%%%%%%%%%%%%%%%%%%%%%%%%%%%%%%%%%%%%%%%
  \begin{tikzpicture}
    \begin{axis} 
		[
	  	ybar,
	  	title=s04,
		  xmin = 0.5,
		  xmax = 12.5,
		  bar width = 6.5,
		]
		\addplot
		[
			color = blue,
			fill = blue,
		]
		table
		[
	  	col sep = semicolon,
	  	x = "Factor",
	  	y = "s04r",
		]
		{branchingFactorHistogram.csv};
		\end{axis}
  \end{tikzpicture}
  &
%%%%%%%%%%%%%%%%%%%%%%%%%%%%%%%%%%%%%%%%%%%%%%%%%%%%%
  \begin{tikzpicture}
    \begin{axis} 
		[
	  	ybar,
	  	title=s05,
		  xmin = 0.5,
		  xmax = 20.0,
		  bar width = 3.8,
		]
		\addplot
		[
			color = blue,
			fill = blue,
		]
		table
		[
	  	col sep = semicolon,
	  	x = "Factor",
	  	y = "s05r",
		]
		{branchingFactorHistogram.csv};
		\end{axis}
  \end{tikzpicture}
  &
%%%%%%%%%%%%%%%%%%%%%%%%%%%%%%%%%%%%%%%%%%%%%%%%%%%%%
  \begin{tikzpicture}
    \begin{axis} 
		[
	  	ybar,
	  	title=s06,
		  xmin = 0.5,
		  xmax = 10,
		  bar width = 8.28,
		]
		\addplot
		[
			color = blue,
			fill = blue,
		]
		table
		[
	  	col sep = semicolon,
	  	x = "Factor",
	  	y = "s06r",
		]
		{branchingFactorHistogram.csv};
		\end{axis}
  \end{tikzpicture}
  &
%%%%%%%%%%%%%%%%%%%%%%%%%%%%%%%%%%%%%%%%%%%%%%%%%%%%%%
  \begin{tikzpicture}
    \begin{axis} 
		[
	  	ybar,
	  	title=s07,
		  xmin = 0.4,
		  xmax = 24.6,
		  bar width = 3.2,
		]
		\addplot
		[
			color = blue,
			fill = blue,
		]
		table
		[
	  	col sep = semicolon,
	  	x = "Factor",
	  	y = "s07r",
		]
		{branchingFactorHistogram.csv};
		\end{axis}
  \end{tikzpicture}%
%%%%%%%%%%%%%%%%%%%%%%%%%%%%%%%%%%%%%%%%%%%%%%%%%%%%%%
\end{tabular}
\caption{Average number of child nodes for every node in generated hierarchies \textbf{with} execution of reassignment procedure. Horizontal axes show the number of children and vertical axes show the number of occurrences (count) in the hierarchies.}\label{fig:branching-factor-histogram-R}
\end{figure*} %5
	
    \section{Benchmarking dataset}
    \label{sec:benchmarking_dataset}
    One of the goals of this article is to provide the research community with a systematic and comprehensive approach for benchmarking OCH methods. To achieve this goal, the benchmarking dataset of 160 hierarchies was created. These hierarchies were chosen from the 1,600 hierarchies analysed in~\Cref{experiments}.
    
    The process of selecting 160 hierarchies to be included in the benchmarking dataset was as follows. For every parameter set-ups out of 16 possibilities~(see \Cref{tab:parameter-configurations}), 100 hierarchies $h_{i}^{s}$ were generated, where
    \begin{itemize}
        \item $s\in\{$\textit{s00, s00r, s01, s01r, s02, s02r, s03, s03r, s04, s04r, s05, s05r, s06, s06r, s07, s07r}$\}$,
        \item $1 \leq i \leq100$.
    \end{itemize}
    Each of these 1,600 hierarchies has been described by a vector
    \begin{equation}
        descr({h_{i}^s}) = (N_{i}^{s}, L_{i}^{s}, D_{i}^{s}, B_{i}^{s}, P_{i}^{s}),
    \end{equation}
    where 
    \begin{itemize}
        \item $N_i^s$ is the number of nodes in the $i$-th hierarchy for parameter set-up $s$,
        \item $L_i^s$ is the number of leaves in the $i$-th hierarchy for parameter set-up $s$,
        \item $D_i^s$ is the depth of the $i$-th hierarchy for parameter set-up $s$,
        \item $B_i^s$ is the breadth of the $i$-th hierarchy for parameter set-up $s$,
        \item $P_i^s$ is the average length of all paths in the $i$-th hierarchy for parameter set-up $s$.
    \end{itemize}
    Additionally, for every parameters set-ups a vector of average values $H_{avg}^s$ has been established
    \begin{equation}
        H_{avg}^s = (\bar{N}^{s}, \bar{L}^{s}, \bar{D}^{s}, \bar{B}^{s}, \bar{P}^{s}),
    \end{equation}
    where $\bar{N}^{s}$, $\bar{L}^{s}$, $\bar{D}^{s}$, $\bar{B}^{s}$, $\bar{P}^{s}$ are values of  $\bar{N}$, $\bar{L}$, $\bar{D}$, $\bar{B}$, $\bar{P}$  from~\Cref{tab:quantitative-hierarchy-analysis,tab:quantitative-hierarchy-analysis-R} for a corresponding $s$.
    
    For all values of parameter $s$ separately, vectors $descr(h_{i}^s)$ and $H_{avg}^s$ have been min-max scaled to range from 0 to 1. In the scaled space, Euclidean distance has been computed between an average vector and all the corresponding hierarchy vectors. Based on the distances the top 10 closest hierarchies have been identified. These hierarchies were published as benchmarking data for a particular parameter set-up. The procedure has been repeated for all parameter set-ups resulting in a collection of 160 hierarchies that are publicly available  (\url{http://kio.pwr.edu.pl/?page\_id=396}).
    
    Quantitative measures for the published dataset are reported in~\Cref{tab:datasets-characteristics}. This table presents results in the same format as in~\Cref{tab:quantitative-hierarchy-analysis,tab:quantitative-hierarchy-analysis-R} with two additional measures:
    \begin{itemize}
        \item $\bar{C}$ -- the average number of immediate children for every internal node (node with at least one non-empty child) in a hierarchy, averaged over all generated hierarchies,
        \item $\bar{I}$ -- the average number of instances per node in a hierarchy, averaged over all generated hierarchies.
    \end{itemize}
    \begin{table*}[h!]
\setlength{\tabcolsep}{3pt}
\begin{center}
\caption{Accumulative characteristics of hierarchies published as benchmarking dataset. Average $\bar{X}$ values together with standard deviation $\sigma_{\bar{X}}$ (or an average of standard deviations $\bar{\sigma}_{\bar{X}}$) are provided.\label{tab:datasets-characteristics}}
\hspace*{-1.1cm}
\begin{tabular}{lrrrrrrrrrrrr}
\multicolumn{1}{c}{\multirow{3}{*}{\textit{Set}}} & \multicolumn{2}{c}{\multirow{2}{*}{\textit{Nodes}}} & \multicolumn{2}{c}{\multirow{2}{*}{\textit{Leaves}}} & \multicolumn{2}{c}{\multirow{2}{*}{\textit{Depth}}} & \multicolumn{2}{c}{\multirow{2}{*}{\textit{Breadth}}} & \multicolumn{2}{c}{\textit{Child per}} & \multicolumn{2}{c}{\textit{Instances}} \\
\multicolumn{1}{c}{} & \multicolumn{2}{c}{} & \multicolumn{2}{c}{} & \multicolumn{2}{c}{} & \multicolumn{2}{c}{} & \multicolumn{2}{c}{\textit{Int. Node}} & \multicolumn{2}{c}{\textit{per Node}} \\
\multicolumn{1}{c}{} & \multicolumn{1}{c}{\textit{$\bar{N}$}} & \multicolumn{1}{c}{\textit{$\sigma_{\bar{N}}$}} & \multicolumn{1}{c}{\textit{$\bar{L}$}} & \multicolumn{1}{c}{\textit{$\sigma_{\bar{L}}$}} & \multicolumn{1}{c}{\textit{$\bar{D}$}} & \multicolumn{1}{c}{\textit{$\sigma_{\bar{D}}$}} & \multicolumn{1}{c}{\textit{$\bar{B}$}} & \multicolumn{1}{c}{\textit{$\bar{\sigma}_{\bar{B}}$}} & \multicolumn{1}{c}{\textit{$\bar{C}$}} & \multicolumn{1}{c}{\textit{$\bar{\sigma}_{\bar{C}}$}} & \multicolumn{1}{c}{\textit{$\bar{I}$}} & \multicolumn{1}{c}{\textit{$\bar{\sigma}_{\bar{I}}$}} \\ \hline
\textit{s00} & 15.10 & 2.60 & 7.00 & 1.56 & 3.90 & 0.32 & 3.09 & 1.69 & 1.75 & 0.65 & 679.08 & 1705.60 \\
\textit{s00r} & 13.90 & 2.23 & 6.70 & 1.16 & 3.70 & 0.48 & 2.98 & 1.58 & 1.81 & 0.63 & 736.32 & 1602.21 \\
\textit{s01} & 109.20 & 27.29 & 32.10 & 8.09 & 13.10 & 1.20 & 7.68 & 4.81 & 1.41 & 0.53 & 98.32 & 522.55 \\
\textit{s01r} & 85.40 & 23.68 & 28.50 & 7.53 & 11.60 & 1.65 & 6.72 & 4.36 & 1.49 & 0.56 & 125.29 & 532.46 \\
\textit{s02} & 592.70 & 116.10 & 225.60 & 36.93 & 12.60 & 0.97 & 43.43 & 43.04 & 1.62 & 1.18 & 17.41 & 215.91 \\
\textit{s02r} & 547.50 & 132.65 & 308.50 & 83.20 & 11.60 & 0.84 & 43.70 & 41.21 & 2.28 & 1.58 & 19.24 & 180.24 \\
\textit{s03} & 50.90 & 8.40 & 21.70 & 4.16 & 6.20 & 0.42 & 7.08 & 5.20 & 1.71 & 0.67 & 201.36 & 707.08 \\
\textit{s03r} & 50.80 & 7.86 & 23.20 & 5.20 & 6.10 & 0.32 & 7.15 & 5.27 & 1.80 & 0.71 & 201.38 & 441.08 \\
\textit{s04} & 3711.80 & 502.46 & 445.50 & 58.94 & 50.80 & 2.04 & 71.93 & 70.96 & 1.14 & 0.30 & 2.74 & 45.35 \\
\textit{s04r} & 2839.10 & 242.45 & 548.70 & 47.38 & 49.50 & 3.14 & 56.21 & 55.77 & 1.24 & 0.40 & 3.55 & 37.76 \\
\textit{s05} & 468.90 & 47.42 & 260.80 & 25.08 & 6.90 & 0.32 & 59.34 & 60.97 & 2.25 & 1.75 & 21.53 & 127.33 \\
\textit{s05r} & 512.50 & 87.08 & 336.90 & 55.25 & 6.50 & 0.53 & 68.32 & 67.81 & 2.93 & 2.15 & 19.99 & 64.46 \\
\textit{s06} & 167.20 & 26.73 & 60.70 & 9.87 & 8.50 & 0.53 & 17.64 & 14.40 & 1.56 & 0.70 & 61.38 & 252.74 \\
\textit{s06r} & 173.60 & 20.49 & 69.50 & 7.46 & 8.30 & 0.48 & 18.73 & 15.11 & 1.66 & 0.71 & 58.37 & 96.44 \\
\textit{s07} & 2221.70 & 242.26 & 1032.20 & 106.40 & 9.10 & 0.32 & 220.53 & 243.58 & 1.87 & 1.36 & 4.56 & 23.28 \\
\textit{s07r} & 2285.50 & 210.33 & 1360.20 & 123.66 & 9.40 & 0.52 & 220.00 & 247.80 & 2.47 & 1.81 & 4.41 & 8.71
\end{tabular}
\end{center}
\end{table*}
    
     The results presented in~\Cref{tab:datasets-characteristics} are very similar to these presented in~\Cref{tab:quantitative-hierarchy-analysis,tab:quantitative-hierarchy-analysis-R} with differences mainly attributed to the stochastic nature of the generator. Similarities in the results indicate that the chosen set of published 160 hierarchies is a representative sample and all the conclusions from~\Cref{experiments} apply to them as well. Specifically, the trends (how the hierarchy statistics change between levels) showed in~\Cref{fig:level-by-level-number-of-instances,fig:level-by-level-number-of-instances-R,fig:level-by-level-hierarchy-width,fig:level-by-level-hierarchy-width-R,fig:level-by-level-number-of-children-per-node,fig:level-by-level-number-of-children-per-node-R,fig:level-by-level-number-of-leaves,fig:level-by-level-number-of-leaves-R,fig:branching-factor-histogram,fig:branching-factor-histogram-R} apply to the published sample as well.
    
    The average number of children per internal node ($\bar{C}$) is usually between 1 and 2, which indicates that the hierarchies are quite narrow (\Cref{tab:datasets-characteristics}). A manual inspection of hierarchies confirmed that nodes with multiple (e.g., 5) children happen but far less often. The reported average number of instances per node ($\bar{I}$) differs significantly  between sets, e.g., 2.74 for \textit{s04} and 679.08 for \textit{s00}. Furthermore, high standard deviation values ($\bar{\sigma}_{\bar{I}}$) indicate high variability in node sizes within hierarchies that might be compensated by the reassignment procedure. 
    
    The reassignment post-processing procedure has the largest impact on the instances per node measure, whereas the differences in other metrics (when hierarchies for reassigned and not reassigned parameter set-ups are compared) are not that significant. They rather stem from differences between samples themselves (separate hierarchies have been chosen for \textit{s00} and for \textit{s00r}). Reassignment procedure acts as a denoising component relocating instances between nodes but leaving hierarchy structures unchanged. This is in line with earlier observations made in this article.
    
    The key summary characteristics of the published hierarchies are presented in~\Cref{tab:datasets-summary}. They are combined for the reassigned and not reassigned variants for the reasons provided in the paragraphs above. The table provides a quick reference for a potential user assisting in a decision of which hierarchies to use in a particular use-case. The decision whether to use the reassigned variant of not should be based on the fact that reassignment denoise the hierarchy, so it is expected to be easier to cluster the points. 
    
    One use-case for the benchmarking dataset might be a development of a new OCH method. If a prototype is being developed, and authors want to evaluate the performance of the method on very high and narrow hierarchies, \textit{s04} or \textit{s04r} should be used. On the other hand, if performance on not high but wider hierarchies is to be validated, \textit{s05} or \textit{set05r} are a better choice. An additional benefit of the benchmarking dataset is that the complexity of a particular hierarchy aspect (width, depth, number of nodes, number of instances per node) can be changed gradually in the series of experiments validating method scalability. For example, to test the method's performance on hierarchies of similar breadth and depth, the initial experiments might be conducted on \textit{s00} or \textit{s00r} which, on average, are 3.9 deep ($\bar{D}$), 3.09 nodes wide ($\bar{B}$), and consist of 15.1 nodes ($\bar{N}$) (for the reassignment variant $\bar{D}$ = 3.7; $\bar{B}$ = 2.98 and $\bar{N}$ = 13.9). The second round of tests can be conducted on \textit{s03} or \textit{s03r} hierarchies that are still characterised by a similar breadth and depth, but in this case, the hierarchies are higher, wider ($\bar{D}$ = 6.2; $\bar{B}$ = 7.08 for \textit{s03} and $\bar{D}$ = 6.1; $\bar{B}$ = 7.15 for \textit{s03r}) and have an increased number of nodes (50.9 for \textit{s03} and 50.8 for \textit{s03r}). 
    
    The established benchmarking dataset provides research communities with a standardised approach to compare their methods. The published hierarchies cover a variety of different hierarchical structures and point distributions allowing for a comprehensive method evaluation but they don't provide all the possible hierarchies. In that case, the generator described in this article can be used in order to generate hierarchies with the desired characteristics. 
    \begin{table*}[h!]
\setlength{\tabcolsep}{3pt}
\begin{center}
\caption{Summary description of hierarchies published as benchmarking dataset.\label{tab:datasets-summary}}
%\hspace*{-4cm}
\begin{tabular}{ll}
% \cline{1-2}
\multicolumn{1}{c}{\textit{Set}} & \multicolumn{1}{c}{\textit{Summary Description}} \\ \cline{1-2}
\multicolumn{1}{|l}{\multirow{3}{*}{\begin{tabular}{@{}l@{}}\textit{s00} \\ \textit{s00r}\end{tabular}}} & \multicolumn{1}{l|}{-- the smallest hierarchical structures} \\
\multicolumn{1}{|l}{} & \multicolumn{1}{l|}{-- similar depth and breadth} \\
\multicolumn{1}{|l}{} & \multicolumn{1}{l|}{-- the highest number of instances per node} \\ \cline{1-2}
\multicolumn{1}{|l}{\multirow{4}{*}{\begin{tabular}{@{}l@{}}\textit{s01} \\ \textit{s01r}\end{tabular}}} & \multicolumn{1}{l|}{-- the most longitudinal hierarchies (high and narrow)} \\
 \multicolumn{1}{|l}{} & \multicolumn{1}{l|}{-- high number of instances per node} \\
 \multicolumn{1}{|l}{} & \multicolumn{1}{l|}{-- low number of nodes} \\
 \multicolumn{1}{|l}{} & \multicolumn{1}{l|}{-- low average number of children per node} \\ \cline{1-2}
\multicolumn{1}{|l}{\multirow{4}{*}{\begin{tabular}{@{}l@{}}\textit{s02} \\ \textit{s02r}\end{tabular}}} & \multicolumn{1}{l|}{-- medium number of nodes} \\
 \multicolumn{1}{|l}{} & \multicolumn{1}{l|}{-- wide hierarchies} \\
 \multicolumn{1}{|l}{} & \multicolumn{1}{l|}{-- low average number of instances per node} \\
 \multicolumn{1}{|l}{} & \multicolumn{1}{l|}{-- medium number of leaves} \\ \cline{1-2}
 \multicolumn{1}{|l}{\textit{s03}} & \multicolumn{1}{l|}{\multirow{2}{*}{-- similar to \textit{s00} and \textit{s00r} but a little larger structures}} \\
\multicolumn{1}{|l}{\textit{s03r}} & \multicolumn{1}{l|}{} \\ \cline{1-2}
\multicolumn{1}{|l}{\multirow{4}{*}{\begin{tabular}{@{}l@{}}\textit{s04} \\ \textit{s04r}\end{tabular}}} & \multicolumn{1}{l|}{-- the highest hierarchies} \\
 \multicolumn{1}{|l}{} & \multicolumn{1}{l|}{-- the largest number of nodes} \\
 \multicolumn{1}{|l}{} & \multicolumn{1}{l|}{-- very narrow} \\
 \multicolumn{1}{|l}{} & \multicolumn{1}{l|}{-- the lowest average number of children per node} \\ \cline{1-2}
\multicolumn{1}{|l}{\multirow{3}{*}{\begin{tabular}{@{}l@{}}\textit{s05} \\ \textit{s05r}\end{tabular}}} & \multicolumn{1}{l|}{-- very wide hierarchies} \\
 \multicolumn{1}{|l}{} & \multicolumn{1}{l|}{-- medium number of nodes and leaves} \\
 \multicolumn{1}{|l}{} & \multicolumn{1}{l|}{-- the largest number of children per node} \\ \cline{1-2}
\multicolumn{1}{|l}{\textit{s06}}  & \multicolumn{1}{l|}{\multirow{2}{*}{-- similar to \textit{s01} and \textit{s01r} but a bit larger structures}} \\ 
\multicolumn{1}{|l}{\textit{s06r}} & \multicolumn{1}{l|}{} \\ \cline{1-2}
\multicolumn{1}{|l}{\multirow{5}{*}{\begin{tabular}{@{}l@{}}\textit{s07} \\ \textit{s07r}\end{tabular}}} & \multicolumn{1}{l|}{-- very large number of nodes} \\
 \multicolumn{1}{|l}{} & \multicolumn{1}{l|}{-- the widest hierarchies} \\
 \multicolumn{1}{|l}{} & \multicolumn{1}{l|}{-- the largest number of leaves} \\
 \multicolumn{1}{|l}{} & \multicolumn{1}{l|}{-- very low average number of instances per node} \\
 \multicolumn{1}{|l}{} & \multicolumn{1}{l|}{-- very high average number of children per node} \\
 \cline{1-2}
\end{tabular}
\end{center}
\end{table*}
    
    \section{Conclusions}
    \label{conclusion}
    In this paper a novel generator of synthetic hierarchical data and a set of benchmarking datasets have been proposed. They aim at providing the research community with a systematic way to benchmark OCH clustering methods and to assist with further OCH development (e.g., development of clustering validation measures). This research has presented a thorough (theoretical and empirical) analysis of the generator that should provide the reader with a comprehensive understanding of how to use it.
    
    The experiments presented in the previous sections highlight both strengths and weaknesses of the proposed generator. A prominent strength is a high range of different tree structures that can be generated and the ability to fine-control these structures using the introduced parameters. Because of that, a wide range of different hierarchy types, often seen in the real-world problems~\cite{ghahramani2010tree}, has been generated and made publicly available\footnote{\url{http://kio.pwr.edu.pl/?page\_id=396}}. 
    
    Published benchmarks and the ability to create more hierarchies using the generator is laying a solid foundation for further development of the concept of Object Cluster Hierarchies. Generated hierarchies can assist not only in clustering methods development and comparison but also in research of OCH quality measures. The latter is important since existing internal and external measures do not fully recognise the differences between OCH and Hierarchical Clustering results. 
    
    From a practical point of view, it is also beneficial that the generator's parameters can be separated into groups, each controlling a different aspect of the hierarchy. Vertical distribution of data is controlled by $\alpha_0$ and $\lambda$, hierarchy width depends on the value of $\gamma$, and $p$, $q$ controls the data specificity. As shown in this article, every parameter set has an interpretation, and its effect on the generated hierarchy is straightforward. This allows for a further fine-tuning towards desired test data. The generation process scales with the number of points to generate, expanding the hierarchy as more elements are generated.
    
    One of the conclusions that emerged from the experiments was that generated hierarchies would display a degree of self-similarity replicating the same general form both vertically and horizontally. We called it as a generation of trailing divisions. Because of that, a few specific areas of the generated hierarchies are not fully controlled. A remedy to this issue is to use a post-processing procedure similar to the reassignment process described in~\Cref{sec:reassignment_post_processing}. This should result in cleaner hierarchies and give the user more control over their overall structure.
    
    In its current form, the generator is limited to a generation of normally-distributed, multidimensional, and uncorrelated real value data. It can be extended to use different kernels leading to different structures of generated hierarchies or generators operating on different types of data.
    
    Furthermore, the self-similar (fractal) nature of the hierarchies suggests a potential for the generator to be described using the language of fractals and especially L-Systems~\cite{prusinkiewicz2013lindenmayer}. Describing the generation process in that form may provide a different (more granular) view over the details of the hierarchical structure. 
	\bibliography{mybibfile}
\end{document}